\documentclass{article}

\usepackage{arxiv}
\usepackage[utf8]{inputenc}
\usepackage[T1]{fontenc}
\usepackage[english]{babel}
\usepackage{amsmath,amssymb,amsfonts}
\usepackage{graphicx}
\usepackage{subcaption}
\usepackage{booktabs}
\usepackage{tabularx}
\usepackage{array}
\usepackage{ragged2e}
\usepackage{adjustbox}
\usepackage{multirow}
\usepackage{pdflscape}
\usepackage{rotating}
\usepackage{algorithm}
\usepackage{algpseudocode}
\usepackage{nicefrac}
\usepackage{microtype}
\usepackage{textcomp}
\usepackage{xcolor}
\usepackage{geometry}
\usepackage[section]{placeins}
\usepackage{float}
\usepackage{titlesec}
\usepackage{etoolbox}
\usepackage{nameref}
\usepackage{natbib}
\usepackage{doi}
\usepackage{lipsum}
\usepackage{url}
\usepackage{hyperref}

\title{HyperPersona: A Multi-Level Hypergraph Framework for Text-Based Automatic Personality Prediction}

\author{
	1$^{st}$ Sina Heydari \\
	Department of Computer Science and Information Technology \\ 
	Institude for Advanced Studies in Basic Sciences (IASBS) \\
	Zanjan, Iran, 45137-66731\\
	\texttt{sinaheydari@iasbs.ac.ir} \\
	\And
	2$^{nd}$ Majid Ramezani \\
	Department of Computer Science and Information Technology \\ 
	Institude for Advanced Studies in Basic Sciences (IASBS) \\
	Zanjan, Iran, 45137-66731\\
	\texttt{ramezani@iasbs.ac.ir} \\
}

\begin{document}
	\maketitle
	\begin{abstract}
		As a modern commodity, language has become a vast repository of socially and psychologically significant traits and concepts, reflecting the ways people encode pattern of thoughts, behaviors, and emotions into words. Text-based Automatic Personality Prediction (APP), seeks to infer personality from linguistic behavior, offering a scalable alternative to traditional psychometric assessments. Although text is inherently hierarchical, with the document-level capturing global features, the sentence-level encoding local semantics, and the word-level providing fine-grained lexical information, most existing approaches rely on shallow, sequential, or single-level representations that ignore the multi-level structure of written language. To address this, we propose HyperPersona, a framework that explicitly models the hierarchical organization of text (document, sentence, and word) through hypergraph structure, where a document and its sentences are represented as hyperedges, and the words are represented as nodes, enabling joint modeling of global, local, and lexical dependencies of text. Followed by a transformer-based graph encoder that learns interactions within and across these linguistic layers, yielding context-sensitive and structurally grounded feature representations for personality prediction. Experiments on the Big Five personality dimensions show that, while relying solely on text, HyperPersona effectively integrates multi-level linguistic cues, achieving superior performance compared to state-of-the-art baselines. These findings underscore the critical role of textual hierarchy in advancing human-like personality inference from natural language.
	\end{abstract}

	
	

	\section{Introduction}\label{sec1}
	
	\label{sec:intro}
	\emph{Personality} is the living current of consciousness through which thoughts, emotions, and habits shape the self. Personality is not a fixed entity, but a dynamic process, an ever-flowing stream where experience and action continuously redefine who we are~\citep{james1890}. Through observable behavioral cues that reflect consistent patterns of thought and emotion, individuals exhibit enduring dispositions such as sociability, impulsivity, or conscientiousness, which together constitute measurable personality traits~\citep{mccrae1999}. In recent decades, Internet-mediated communication (encompassing instant messaging platforms, social networks, email systems, online forums, and collaborative digital environments) has expanded at an unprecedented rate. These communication forms, driven by technological advancement and accelerated by global disruptions, have become deeply embedded in human social and professional life~\citep{castells2011}. Understanding individuals’ personalities through their digital interactions offers valuable insights into cognitive tendencies, emotional responses, and behavioral patterns across diverse contexts~\citep{golbeck2011}. Such understanding can enhance interpersonal communication, foster organizational harmony, and improve social and commercial relationships; consequently, personality prediction holds broad potential in numerous domains, including recommender systems, recruitment and talent acquisition, digital marketing, social network analysis, psychological counseling, and human resource management, among others~\citep{feizi2022}.
	
	The study of personality has long been central to psychology, seeking to explain stable behavioral patterns and inter-individual differences; foundational frameworks such as the Big Five model conceptualize personality as comprising five broad dimensions: \emph{Openness (O), Conscientiousness (C), Extraversion (E), Agreeableness(A), and Neuroticism (N)}~\citep{mccrae1999}. Traditional assessment tools, notably the Big Five Inventory (BFI)~\citep{bfi} and the Revised NEO Personality Inventory (NEO-PI-R)~\citep{costa2008}, have provided reliable measures of these traits; however, they are constrained by practical limitations such as reliance on self-reporting, susceptibility to social desirability bias, and limited scalability for large populations~\citep{costa1992, digman1990, defruyt2009, vicentini2025}. The emergence of large-scale online communication has opened new opportunities to overcome these limitations. Through \emph{Automatic Personality Prediction (APP)}, which is the computational assessment of personality traits based on digital behavioral data; and text-based APP, in particular, leverages natural language as a mirror of psychological processes, enabling personality inference from linguistic patterns within human communication~\citep{feizi2022}. Early computational studies relied heavily on handcrafted lexical and psycholinguistic features (such as word usage, emotional tone, or syntactic complexity) combined with traditional machine learning techniques like SVM and logistic regression~\citep{pennebaker2001, argamon2005, schwartz2013}. Although these approaches established the feasibility of computational personality inference, they were limited by their dependency on shallow linguistic features and weak generalization to diverse domains~\citep{stachl2020, feizi2022}. With the advent of deep learning, a new era in APP research was introduced; neural architectures, such as convolutional and recurrent networks, enabled models to automatically learn distributed text representations, reducing the need for manual feature engineering~\citep{majumder2017, lynn2020, gim2018}. Later on, transformer-based encoders (e.g., BERT~\citep{bert}, RoBERTa~\citep{roberta}, and XLNet~\citep{xlnet}) revolutionized the field by modeling long-range contextual dependencies through self-attention mechanisms~\citep{attention}. Despite the revolutionary outcome of significant improvement in predictive performance, yet the transformer-based approaches often treated text as a flat sequence, neglecting the hierarchical structure that naturally spans words, sentences, and documents~\citep{mehta2020, christian2021}. More recently, graph-based and hybrid models have emerged to address this limitation by explicitly encoding semantic and syntactic dependencies, offering greater structural interpretability and showing promising aspects in modeling hierarchical relations of text, while transformer-based architectures continue to dominate in contextual understanding~\citep{wu2021, wang2024}.
	
	However, these paradigms face significant shortcomings, one which being the lack of a framework that models the hierarchical nature of language and captures its local and global aspects, which has motivated a new line of research in hierarchical text modeling that integrates multi-level representations for more comprehensive and interpretable personality prediction~\citep{yang2021ps, ramezani2022oe, zhu2024}. As suggested by Mikolov et al.~\cite{mikolov2013}, Kiros et al.~\cite{kiros2015}, and Le et al.~\cite{le2014}, the \emph{document-level} captures global features, the \emph{sentence-level} encodes local semantics, and the \emph{word-level} contains fine-grained lexical information. To this end, we propose \emph{HyperPersona}, a multi-level hypergraph framework for text-based personality prediction. By exposing text at three interconnected linguistic levels (\emph{document}, \emph{sentence}, and \emph{word}), the model captures lexical, syntactic, and semantic dependencies. HyperPersona operates at four phases of architecture, including: \emph{preprocessing},\emph{text representation}, \emph{representation learning} using a transformer-based graph encoder, and \emph{personality classification}. The principal contribution of this study lies in presenting a unified hypergraph framework that models the hierarchy of language. HyperPersona encodes higher-order dependencies through hyperedges, embedding linguistic structure directly within the graph topology. It further integrates hierarchical representation with transformer-based message passing to achieve scalable, end-to-end learning. To the best of our knowledge, this is the first framework that applies a multi-level structure to capture knowledge from different levels of the input text for automatic personality prediction. This is achieved by using a \emph{hypergraph} to model the multi-level organization of the knowledge represented in the text. The proposed framework (HyperPersona), is based on the Big Five personality model and is trained and evaluated using the Essays dataset. This study sought to address the following research questions:
	\begin{itemize}
		\item \textbf{RQ1.} Does modeling text through a multi-level hypergraph that captures document, sentence, and word-level dependencies improve automatic personality prediction?
		\item \textbf{RQ2.} How does the proposed hypergraph transformer framework perform in comparison with established text classification baselines in predicting personality traits from textual data?
		\item \textbf{RQ3.} How does processing each level of textual hierarchy (i.e. document, sentence and word) affect the model's predictive accuracy?
		\item \textbf{RQ4.} To what extent can a transformer-based model operating over hypergraph representations capture personality-relevant linguistic knowledge across multiple levels of abstraction?
	\end{itemize}
	
	This paper is organized as follows: \nameref{sec:relatedworks} section provides a concise overview of recent developments in text-based APP, categorized into Single-Level~\ref{subsec:singlelevel} and Multi-Level~\ref{subsec:multilevel} approaches. The Methodology section~\ref{sec:methodology} outlines the approach used to develop HyperPersona; starting with the Dataset~\ref{subsec:dataset} description employed in this study, followed by a detailed explanation of how HyperPersona represents textual structures and learns their representations for personality prediction. The Results~\ref{sec:results} section presents the experimental findings, while the Discussion~\ref{sec:discussion} section interprets these results, discusses their implications, and addresses the research questions. Finally, the Conclusion~\ref{sec:conclusion} section summarizes the key findings and outlines potential directions for future research.
	
	\section{Related Works}
	\label{sec:relatedworks}
	Personality research originated in psychology through self-report questionnaires such as the BFI and NEO-PI-R, which, despite their reliability, suffer from limitations including time demands, social desirability bias, and restricted applicability~\citep{costa1992, digman1990, defruyt2009, vicentini2025}. With the rise of large-scale digital communication, computational approaches to APP emerged, initially relying on handcrafted linguistic features and statistical learning~\citep{stachl2020, feizi2022}. These were later surpassed by deep learning models that learned representations directly from raw text, offering significant improvements but treated language as flat sequences~\citep{mehta2020}. Recent efforts have introduced transformers and graph-based methods to capture semantic and syntactic dependencies, thereby offering structural interpretability~\citep{yang2021ps}. However, conventional graphs remain limited in representing the hierarchical organization of text across words, sentences, and documents~\citep{cai2020}. At the same time, pre-trained transformers have delivered state-of-the-art results with minimal preprocessing, yet they too often ignore explicit structural constraints~\citep{christian2021}.
	
	Reflecting its hierarchical nature, we conceptualize written language as a hierarchical structure with three levels: word, sentence, and document. Despite the central role of textual hierarchy in encoding semantics, limited attention has been directed toward modeling personality prediction within a multi-level textual framework. Accordingly, this section provides an overview of existing approaches, categorized into Single-Level~\ref{subsec:singlelevel} and Multi-Level~\ref{subsec:multilevel} text-based approaches.
	
	\subsection{Single-Level}
	\label{subsec:singlelevel}
	In this section, we review single-level text-based methods, which predict personality by treating text as a flat sequence and focusing on modeling only one level of the text hierarchy (either document, sentence, or word) in isolation. Generally, these approaches rely on linguistic features~\citep{feizi2022} or neural encoders~\citep{mehta2020} that capture either local or global semantic and stylistic patterns. However, they overlook the hierarchical organization of language into documents, sentences, and words, collapsing diverse textual knowledge into a unified representation and thereby constraining the capacity to capture nuanced multi-level dependencies.
	
	Early works exemplify the dominance of the \emph{document-level} paradigm. Ong et al.~\cite{ong2017} aggregated a user’s tweets into a single representation, treating each user as a text document, and trained SVMs and XGBoost models on linguistic features. Similarly, Shen et al.~\cite{shen2013}, collapsed the entire body of a user’s emails into a single unit of text (document) for personality inference. In both cases, the choice of document-level integration simplified personality modeling by considering an entire user profile as a block of text, at the expense of overlooking sentence and word-level structures. Skowron et al.~\cite{skowron2016} also adopted the document-level approach, where heterogeneous social media modalities and metadata were combined into a single holistic user profile representation. Ramezani et al.~\cite{ramezani2022text} integrated all text into a document-level structure, which was represented by a knowledge graph, and introduced graph attention networks to capture semantic relations among concepts. Using knowledge graph-enhanced semantic connections, the reliance on a document-level approach prevented the explicit modeling of hierarchical dependencies across different textual units. The rise of transformer-based encoders further solidified this trend.

	
	Jayaraman et al.~\cite{jayaraman2024} processed user data as a continuous flat sequence, making their work a document-level approach. They further employed XLNet, which models bidirectional context through relative positional encoding. Christian et al.\cite{christian2021} ultimately processed user data as a continuous flat sequence, which is also considered a document-level approach, and integrated BERT, RoBERTa, and XLNet with sentiment and emotion features to predict personality from text. Hao et al.~\cite{hao2023} integrated contextual and sentiment knowledge, fusing them into a single global representation at the document-level. They utilized these documents to enrich transformer encoders with sentiment knowledge from SenticNet6. Kamalesh et al.~\cite{kamalesh2022} introduced a pipeline that operates at the document-level, reducing personality modeling to a flat input-output mapping. Then they introduced a Binary-Partitioning Transformer with TF-IGM for discriminative feature extraction. Ganesan et al.~\cite{ganesan2023} systematically evaluated GPT-3 for zero-shot personality estimation at the document-level to determine whether a large language model can infer Big Five traits directly from user-generated text without task-specific fine-tuning. They aggregated each participant’s social media posts into a single document and prompted GPT-3 with carefully designed instructions to output trait scores.
	
	A smaller subset of research has employed \emph{sentence-} or \emph{word-level} approaches. Naz et al.~\cite{naz2025using} focused on sentence-level by employing Bi-LSTM and DistilBERT to obtain sentence embeddings, which were then aggregated for predicting only the openness trait, thereby restricting the model from capturing inter-sentence dependencies or higher-order structures. Similarly, Leonardi et al.~\cite{leonardi2020} employed a multilingual transformer framework to embed Facebook posts at the sentence-level, which addressed cross-lingual polysemy but ultimately reduced each post to a single embedding vector. Kazameini et al.~\cite{kazameini2020} introduced a word-level personality prediction framework that combines contextual BERT embeddings with an ensemble of bagged SVMs. Their approach preserves token-level representations and aggregates them through bootstrap sampling, allowing for a fine-grained capture of lexical cues. Alsini et al.~\cite{alsini2024} proposed a word-level approach for predicting the Agreeableness trait, to assess whether a person is emotional or rational, using deep learning on distributed word representations. They extracted Word2Vec and GloVe embeddings from the MBTI dataset to encode each token’s semantic context and trained LSTM and Bi-LSTM networks to model sequential dependencies among these word vectors.
	
	Overall, single-level approaches demonstrate distinct advantages. Document-level methods aggregate an individual’s entire text profile, yielding stable global representations and simplifying the modeling pipeline while reducing computational cost. Sentence-level methods extend this by modeling local syntactic and semantic structures, which is valuable when personality markers emerge from compositional phrases or clause-level patterns. Word-level models capture fine-grained lexical cues and psycholinguistic features, enabling direct exploitation of trait-relevant vocabulary. Despite these strengths, each level carries notable limitations. Document-level aggregation, although holistic, obscures subtle contextual variations and reduces interpretability by flattening diverse linguistic features. Sentence-level approaches, while sensitive to local dependencies, overlook inter-sentence coherence and broader narrative flow. Word-level modeling often overlooks the influence of syntax and discourse, resulting in fragmented representations and a loss of higher-order context. Across all three levels, hierarchical dependencies remain unmodeled, and language is ultimately treated as a homogeneous sequence of words.
	
	As shown in Table~\ref{tab:singleleveltable}, Single-level methods thus offer a computationally efficient foundation for APP, emphasizing either local or global features in isolation. However, they fall short of jointly capturing hierarchical linguistic cues. Despite their historical significance, such approaches remain inadequate for modeling the multi-layered linguistic structures through which personality is conveyed.
	
	\begin{sidewaystable}[htbp]
		\centering
		\caption{Tabular view of single-level text-based approaches}
		\label{tab:singleleveltable}
		\begin{tabular}{p{3cm} p{1.5cm} p{3cm} p{2cm} p{1.5cm} p{2cm} p{0.6cm}}
			\toprule
			Name & Processing Level & Methods & Dataset & Personality Model & Context & Year \\ \midrule
			Ong et al.~\cite{ong2017} & document & SVM, XGBoost & Twitter  & Big Five & Social media text & 2017 \\
			Shen et al.~\cite{shen2013} & document & Joint, sequential, survival generative models & Enron Email Corpus & Big Five & Email & 2013 \\
			Skowron et al.~\cite{skowron2016} & document & Multimodal Personality Regressors & Social media\footnotemark & Big Five & Social media profiles & 2016 \\
			Jayaraman et al.~\cite{jayaraman2024} & document & XLNet & Essays dataset & Big Five & Essays & 2024 \\
			Christian et al.~\cite{christian2021} & document & Model averaging (BERT, RoBERTa, and XLNet) & Social media\footnotemark[\value{footnote}] & Big Five & Social media posts & 2021 \\
			Hao et al.~\cite{hao2023} & document & Pre-trained Language Models, SenticNet6 & Weibo & Big Five & Social media text & 2023 \\
			Kamalesh et al.~\cite{kamalesh2022} & document & BERT, LSTM & Social media\footnotemark[\value{footnote}] & Big Five & Facebook status & 2022 \\
			Ganesan et al.~\cite{ganesan2023} & doucment & GPT-3, WT-LEX, MFC (Most Frequent Class) & Social media\footnotemark[\value{footnote}] & Big Five & Social media posts & 2023 \\
			Naz et al.~\cite{naz2025using} & sentence & Bi-LSTM & Essays dataset & Openness trait & Essays & 2025 \\
			Leonardi et al.~\cite{leonardi2020} & sentence & Transformer-based deep neural network & Social media\footnotemark[\value{footnote}] & Big Five & Facebook posts & 2020 \\
			Kazameini et al.~\cite{kazameini2020} & word & Bagged SVM, BERT & Essays dataset & Big Five & Social media and essay text & 2020 \\
			Alsini et al.~\cite{alsini2024} & word & Transformer based models, Bi-LSTM & MBTI dataset & Agreeableness trait & MBTI dataset text & 2024 \\
			\bottomrule
		\end{tabular}
	\end{sidewaystable}
	
	\subsection{Multi-Level}
	\label{subsec:multilevel}
	In this section, we review multi level methods that extend personality prediction by modeling text structure across multiple levels of the textual hierarchy, including document, sentence, and word-levels. By explicitly representing relationships between different text levels and linguistic features, these approaches capture richer contextual and structural cues, enabling more nuanced personality inference.
	
	
	Zhu et al.~\cite{zhu2024} integrated word and document-level information by constructing a heterogeneous personality graph for each user. They augmented this graph with LIWC-based psycholinguistic knowledge at the word-level and employed semi-supervised noise-invariant learning to improve generalization. This design allowed the model to connect fine-grained psycholinguistic cues to broader document-level relational reasoning. Zhu et al.~\cite{zhu2022} pursued a similar integration of word and document-level knowledge. They built personality lexicon-based representations, transformed them into heterogeneous word graphs, and embedded them into document-level structures, thereby explicitly linking local lexical cues with global personality expressions. Johnson et al.~\cite{johnson2023ov} combined word and document-level modeling through a knowledge graph enriched with DBpedia, ontology data, emotion lexicons, and psycholinguistic features. Their method first learned aspect-aware word embeddings, then integrated them into a document-level graph for trait detection. This multi-level pipeline enabled them to leverage psychologically grounded word-level features while still allowing for graph-based reasoning across entire documents. Kerz et al.~\cite{kerz2022} also exploited word and document-level integration by combining explicit psycholinguistic feature distributions (word-level) with contextualized representations learned through BERT and BiLSTM (document-level). Their approach demonstrated how handcrafted linguistic features can be synergistically combined with deep encoders to enhance personality prediction.
	
	Salahat et al.~\cite{salahat2023} similarly targeted word and sentence-level by leveraging BERT embeddings for contextualized word and sentence representations and fed them into CNN layers for feature extraction. Their hybrid pipeline demonstrated how pre-trained language models (PLMs) embeddings can be stacked with hierarchical convolutional reasoning to exploit multiple levels simultaneously. Bama et al.~\cite{bama2025} integrated word and sentence-level embeddings in a hierarchical transformer with label attention mechanisms. Their model selectively aligned features to specific trait labels, effectively linking multi-level linguistic knowledge to psychological categories for MBTI classification. Ramezani et al.\cite{ramezani2022oe} advanced a more comprehensive framework that explicitly integrated lexical, semantic, and sequential features. Their ensemble combined term-frequency vectorization (lexical), ontology-based and enriched ontology models (semantic), latent semantic analysis (document-level semantic abstraction), and BiLSTM-based deep learning (sequential word-level). These base models were combined into a meta-model using a hierarchical attention network (HAN), which facilitated reasoning across multiple levels of abstraction. This ensemble demonstrates how multiple modeling strategies can be integrated into a unified, multi-level pipeline.
	
	Yang et al.~\cite{yang2023} addressed sentence and document-level interactions in their model. They integrated hierarchical attention networks (HANs) with pre-trained language models, allowing sentence-level interactions to inform document-level representations. Importantly, their framework incorporated psychological theory, positioning multi-level modeling not only as a computational choice but also as a way to align with established psychological constructs. Yang et al.~\cite{yang2021} proposed Transformer-MD, which combined sentence and document-level modeling. By employing Transformer-XL memory tokens, their approach captured dependencies across multiple short user-generated posts at the document-level while retaining sentence-level contextual modeling within individual posts. This dual integration addressed the fragmented nature of social media text. Liu et al.~\cite{liu2016} developed a compositional and language-independent model for personality trait recognition that combined word and sentence-level features. They encoded tweets in multiple languages using distributed embeddings and aggregated them via a hierarchical composition function to predict Big Five traits. By leveraging both local lexical cues and sentence-level structure, the model could generalize across languages and short-text social media settings, highlighting the value of multi-level representations in low-resource scenarios. Lynn et al.~\cite{lynn2020} proposed a hierarchical approach for user personality prediction by modeling word and message representations. They employed attention mechanisms at the message-level to capture the relative importance of individual posts within a user’s social media history. This hierarchical structure allowed the model to integrate fine-grained lexical features with broader user-level behavioral patterns, improving Big Five trait prediction. Tirotta et al.~\cite{tirotta2023} introduced a multi-level framework that integrates sentence and document-level embeddings for personality prediction. They extracted sentence embeddings from short social media posts using contextual language models and then aggregated these into document-level representations to infer Big Five traits. This combination allowed the model to capture both fine-grained linguistic patterns and cross-sentence dependencies.
	
	As can be seen in Table~\ref{tab:multileveltable}, taken together, these studies show that multi-level approaches systematically capture knowledge from words, sentences, and documents rather than collapsing text into a flat representation. Their strength lies in capturing hierarchical dependencies that reflect how language is naturally structured, resulting in improved contextual awareness and interpretability compared to single-level methods; however, these methods also come with notable challenges: increased computational complexity, reliance on external resources (such as lexicons or ontologies), and difficulty in balancing contributions from different levels without introducing redundancy.
	
	\begin{sidewaystable}[htbp]
		\centering
		\caption{Tabular view of multi-level text-based approaches}
		\label{tab:multileveltable}
		\begin{tabular}{p{3cm} p{1.5cm} p{3cm} p{2cm} p{1.5cm} p{2cm} p{0.6cm}}
			\toprule
			Name & Processing Level & Models & Dataset & Personality Model & Context & Year \\ \midrule
			Zhu et al.~\cite{zhu2024} & word,\newline document & Graph Convolutional Networks (GCNs) & MyPersonality, YouTube, PAN2015 & MBTI & User data & 2024 \\
			Zhu et al.~\cite{zhu2022} & word,\newline document & Heterogeneous Message Passing Graph Neural Network (HPMN), & Essays dataset, psycholinguistic lexicon & Big Five & lexicon-guided Essays dataset text & 2022 \\
			Johnson et al.~\cite{johnson2023ov} & word,\newline document & CNN, RNN, LSTM, BiLSTM & Social media\footnotemark & Big Five & Social media text  & 2023 \\
			Kerz et al.~\cite{kerz2022} & word,\newline document & BERT, BiLSTM & Essays dataset & Big Five & Essays dataset text & 2023 \\
			Salahat et al.~\cite{salahat2023} & word,\newline sentence & Convolutional Neural Networks (CNNs) & Social media\footnotemark[\value{footnote}] & Big Five & Social media posts & 2023 \\
			Lynn et al.~\cite{lynn2020} & word,\newline sentence & Hierarchical Attention Network (HAN) & Social media\footnotemark[\value{footnote}] & Big Five & Social media posts (multiple per user) & 2020 \\
			Bama et al.~\cite{bama2025} & word,\newline sentence & Hierarchical Transformer network with Label Attention for Personality Prediction (HT-LA-PP) & MBTI dataset (social media\footnotemark[\value{footnote}]) & MBTI & Social media posts & 2025 \\
			Ramezani et al.~\cite{ramezani2022oe} & word,\newline document & Hierarchical Attention Network (HAN) & Essays dataset & Big Five & Essays dataset text & 2022 \\
			Yang et al.~\cite{yang2023} & sentence,\newline document & Hierarchical Attention Network (HAN) & Essays dataset & Big Five & Essays dataset text & 2023 \\
			Yang et al.~\cite{yang2021} & sentence,\newline document & Transformer-MD, Dimension Attention & Social media\footnotemark[\value{footnote}] & Big Five & Social media posts & 2021 \\
			Liu et al.~\cite{liu2016} & word,\newline sentence & Bi-GRU & Twitter & Big Five & Tweets in multiple languages & 2016 \\
			Tirotta et al.~\cite{tirotta2023} & sentence,\newline document & SBERT & Twitter, MNLI (as auxiliary) & Big Five & Short posts and tweets & 2023 \\
			\bottomrule
		\end{tabular}
	\end{sidewaystable}
	
	\subsection{Research Gaps}
	Despite the range of methods reviewed in the preceding categories (single-level approaches~\ref{subsec:singlelevel} and multi-level approaches~\ref{subsec:multilevel}), several critical gaps remain. First, although some studies are categorized as multi-level, no existing framework fully exploits the hierarchical nature of text across all syntactic levels (namely, word, sentence, and document) within a unified model. Second, hypergraph representations, which naturally capture multi-level dependencies by extending beyond pairwise relations and encoding hierarchical structures, have received little to no attention in APP~\citep{antelmi2023, bendersky2012}.
	
	\section{Methodology}
	
	\subsection{Dataset}
	\label{subsec:dataset}
	The textual data used in this work was extracted from the Essays dataset, introduced by Pennebaker and King~\cite{linguisticstyles}. This dataset comprises over 2,400 self-authored essays, labeled with Big Five personality scores from self-assessments. Its free-form structure and direct personality annotations have established it as a benchmark for evaluating text-based models of APP. Developed as part of a psychological research program, the dataset aims to examine whether language use reflects stable personality traits. It employs a word-based, computerized text analysis approach and was one of the first empirical efforts to link natural language to psychological characteristics defined by the Big Five personality traits. The Essays dataset is widely used in personality prediction, natural language processing, and computational psycholinguistics. It contains a total of 2,468 essays written by over 1,200 undergraduate students, who each completed four assignments across a semester. The students wrote on various unrestricted topics, including stream-of-consciousness writing and personal reflections. Each essay ranged from 500 to 800 words, with no strict constraints on length, grammar, or style.
	
\begin{figure}[htbp]
		\centering
		\includegraphics[width=1\textwidth, keepaspectratio]{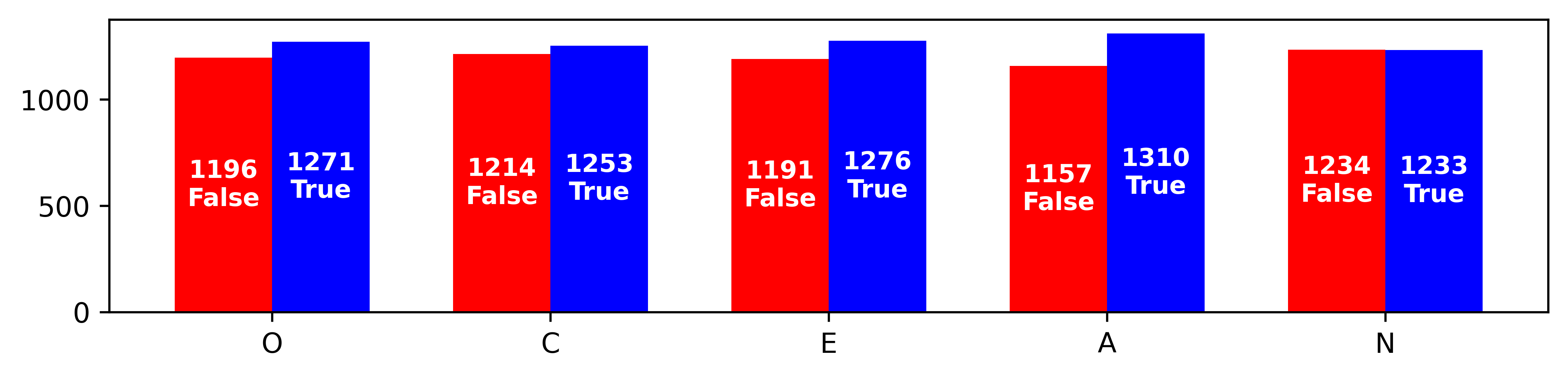}
		\caption{Label distribution for Big Five personality traits in the Essays dataset}
		\label{fig:labeldist}
	\end{figure}

	Figure~\ref{fig:labeldist} illustrates the distribution of binary personality labels in the dataset, where each Big Five trait (OCEAN) is labeled as either True (0) or False (1). The distribution is relatively balanced across traits, which is ideal for classification tasks. Openness has 1,196 samples labeled false and 1,271 true, indicating a slightly higher presence of open-minded individuals. Conscientiousness shows 1,214 false and 1,253 true samples, indicating a mild skew toward the presence of the trait. Extraversion is true in 1,276 cases and false in 1,191, suggesting a majority of extroverted participants. Agreeableness shows the largest gap, with 1,310 true labels and 1,157 false labels, indicating a generally cooperative and trusting population. Neuroticism is almost perfectly balanced, with 1,234 false and 1,233 true cases, offering a symmetric distribution. This balance across traits helps reduce bias and supports reliable training and evaluation of personality prediction models.
	
\begin{figure}[htbp]
		\centering
		\begin{subfigure}[b]{0.49\textwidth}
			\includegraphics[width=\textwidth]{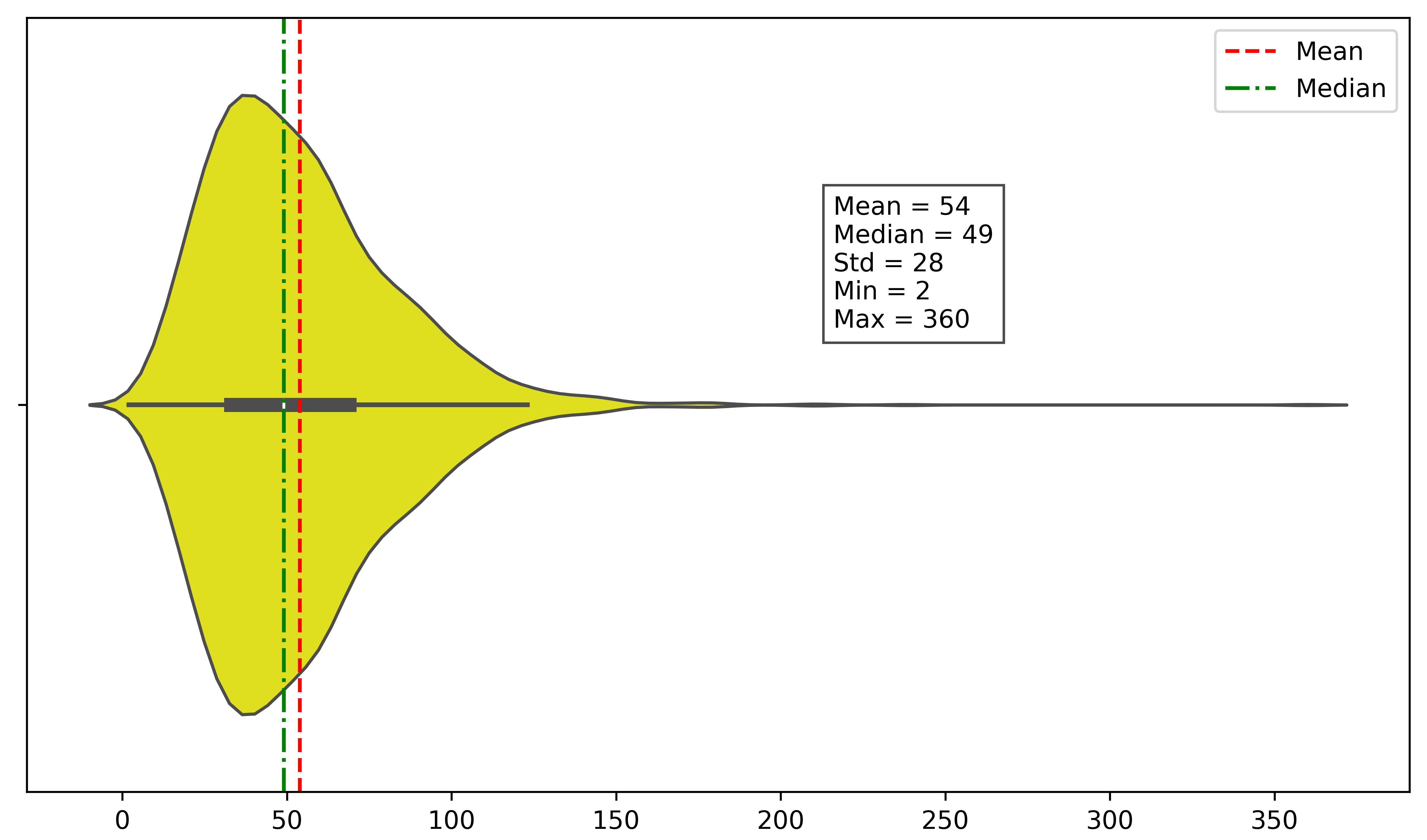}
			\caption{Sentence Distribution}
			\label{fig:ev-a}
		\end{subfigure}
		\hfill
		\begin{subfigure}[b]{0.49\textwidth}
			\includegraphics[width=\textwidth]{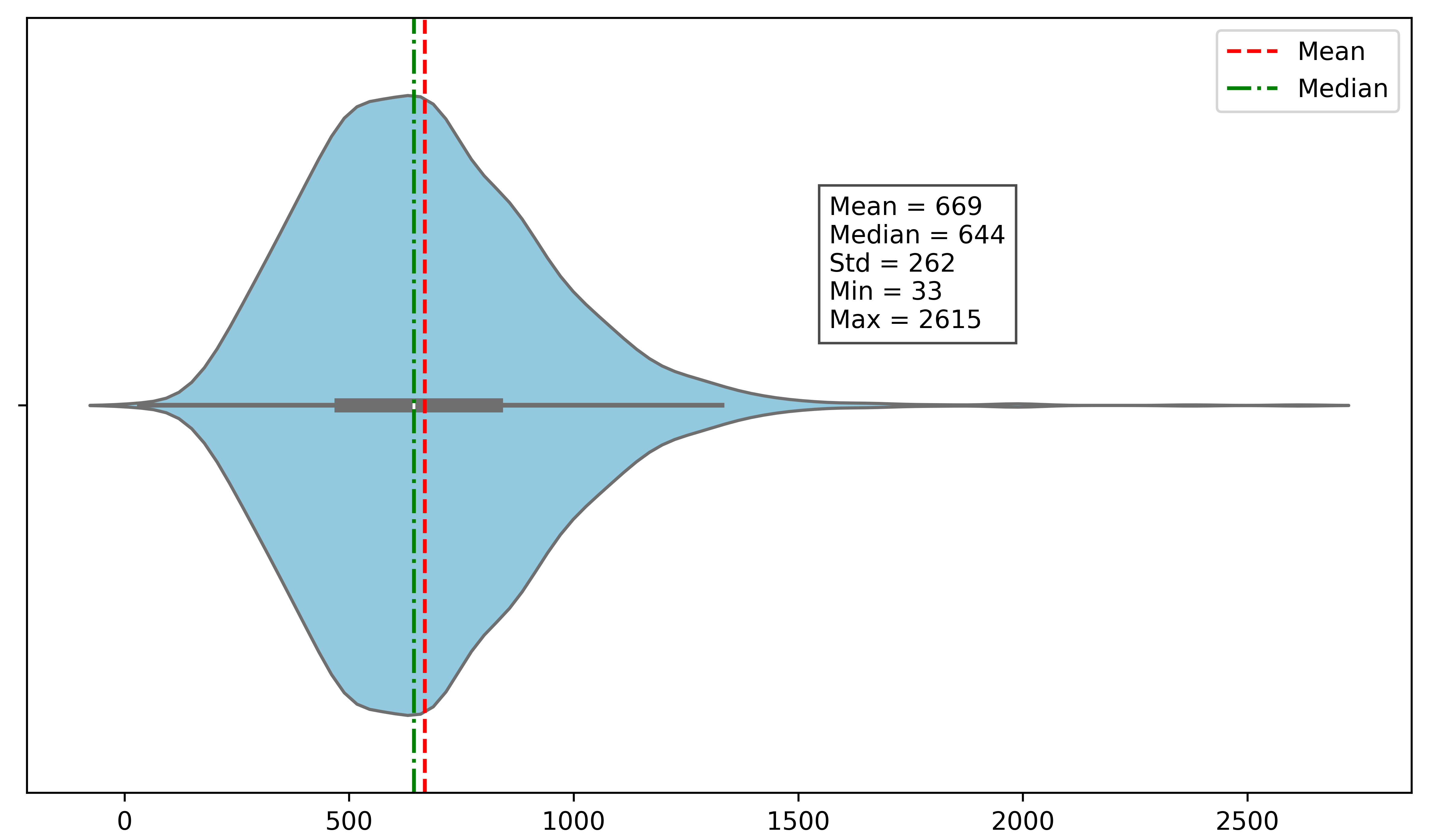}
			\caption{Word Distribution}
			\label{fig:ev-b}
		\end{subfigure}
		\caption{Sentence(a) and Word(b) Distribution in the Essays Dataset}
		\label{fig:essaysviolinplot}
	\end{figure}

	Figure~\ref{fig:essaysviolinplot} depicts the distributions of documents in terms of word and sentence counts. The sentence count distribution (Figure~\ref{fig:ev-a}) is right-skewed, with a mean of 54 and a median of 49 sentences per document, and a standard deviation (Std) of 28. The minimum sentence count is two, and the maximum is 360, indicating outlier documents. Most distributions lie between 30 and 70 sentences, and the slight skew shows the mean is higher than the median. The word count distribution (Figure~\ref{fig:ev-b}) is also right-skewed, with a mean of 669 and a median of 644 words per document, and a standard deviation of 262. The minimum word count is 33, and the maximum is 2,615, highlighting some significantly longer documents that contribute to the skew. The relatively small difference between the mean and median suggests mild outliers. The density of word counts is concentrated between 500 and 800.

	\subsection{Proposed Method}
	\label{sec:methodology}
\begin{figure}[htbp]
		\centering
		\includegraphics[width=\textwidth, height=0.92\textheight, keepaspectratio]{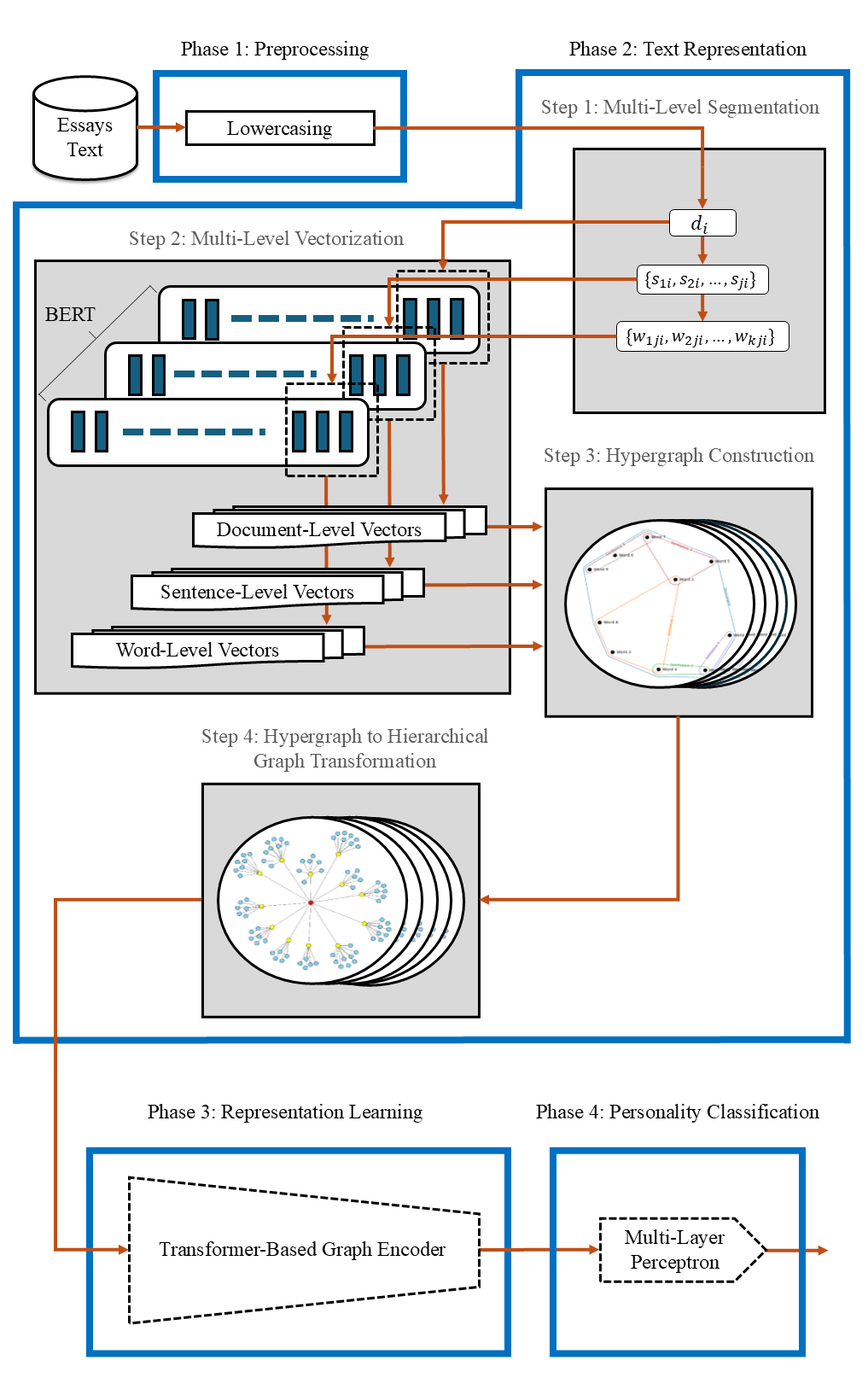}
		\caption{The overall architecture of HyperPersona}
		\label{fig:overall structure}
	\end{figure}

	We propose HyperPersona, a text-driven multi-level hypergraph and hierarchical graph framework which employs a transformer-based graph encoder to predict personality from textual sources. As illustrated in Figure~\ref{fig:overall structure}, our pipeline consists of four main phases: \emph{\nameref{subsubsec:preprocessing}, \nameref{subsubsec:text representation}, \nameref{subsubsec:rep-learning-dim-reduction}, \nameref{subsubsec:personality-classification}}, which will be described in detail.
	
	\subsubsection{Phase 1: Preprocessing}
	\label{subsubsec:preprocessing}
	We applied minimal preprocessing to preserve the linguistic richness. Specifically, we performed lower casing to ensure consistency, while deliberately avoiding other common preprocessing techniques. This decision aligns with the transformer-based nature of our vectorization method (in~\ref{paragraph:multi-level vectorization}), which has enabled us to work with raw text and benefit from the retention of both syntactic and semantic information.
	
	\subsubsection{Phase 2: Text Representation}
	\label{subsubsec:text representation}
	The phase 2 of our pipeline includes four core steps: \emph{\nameref{paragraph:multi-level segmentation}, \nameref{paragraph:multi-level vectorization}, \nameref{paragraph:hypergraph-construction}, \nameref{paragraph:hypergraph to hierarchical graph}}
	which will be explained in detal.
	
\begin{figure}[htbp]
		\centering
		\includegraphics[width=0.96\textwidth, keepaspectratio]{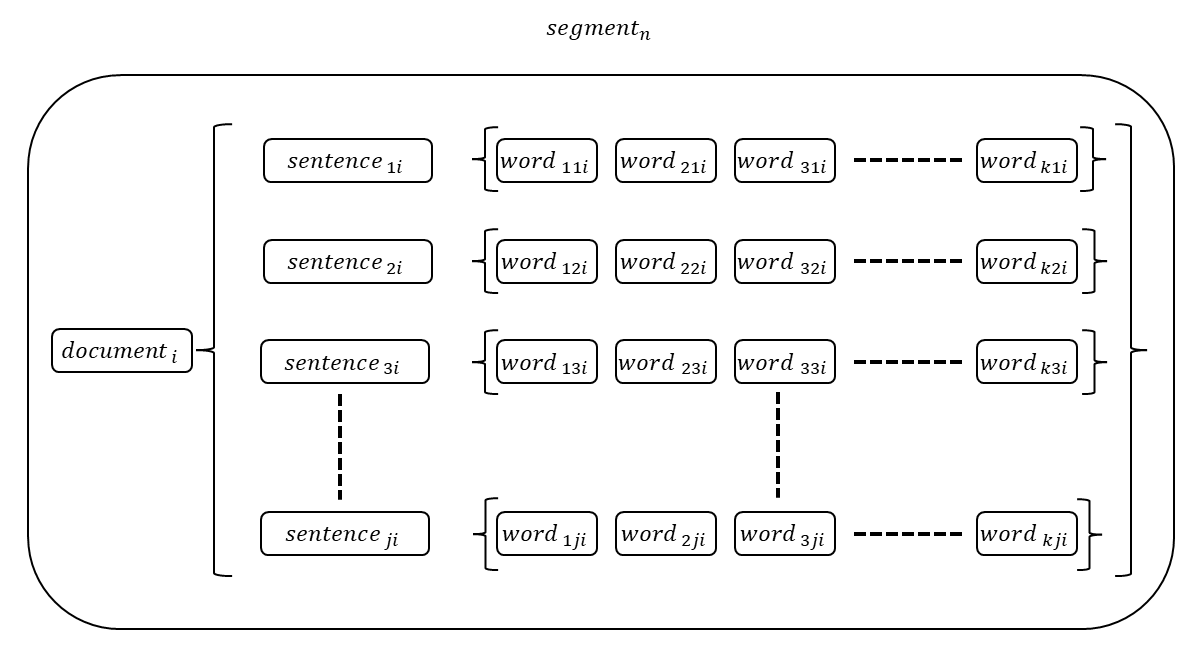}
		\caption{Multi-level segmentation of a document}
		\label{fig:multilevelsegmentation}
	\end{figure}

	\paragraph{Step 1: Multi-level Segmentation}
	\label{paragraph:multi-level segmentation}
	As our multi-level strategy aims to interpret the hierarchical nature of text for the machine, we first needed to break a document down into its syntactic components. Each document is decomposed into a hierarchical sequence of segments modeling its syntactic structure. Figure~\ref{fig:multilevelsegmentation}, depicts the nested structure each document is segmented in, where $document_i$,  $1 \leq i \leq \mid \text{Essays  dataset}\mid$, forms the document-level segment, $sentence_{ji}$, $1 \leq j \leq \text{number of sentences in} document_i$, form the sentence-level segments, and $word_{kji}$, $1 \leq k \leq \text{number of words in } sentence_ji$ form the word-level segment contained within each sentence. The Eq.~\ref{eq:segcorp} shows our corpus, where each document is transformed into a nested segment:
	\begin{center}
		\begin{equation}
			\begin{aligned}
				segmented\_corpus &= \lbrace segment_1, segmet_2, segment_3, \dots, segment_n \rbrace, \\
				segment_n &= \Bigg( document_i, \Big\{ 
				sentence_{1i}, \{ word_{11i}, word_{21i}, \dots, word_{k1i} \}, \\
				&\quad sentence_{2i}, \{ word_{12i}, word_{22i}, \dots, word_{k2i} \}, \dots, \\
				&\quad sentence_{ji}, \{ word_{1ji}, word_{2ji}, \dots, word_{kji} \} 
				\Big\} \Bigg), \\
			\end{aligned}
			\label{eq:segcorp}
		\end{equation}
	\end{center}
	where $i = n$, we decompose $document_i$ into a set of sentences $\lbrace sentence_{1i}, sentence_{2i}, sentence_{3i}, \dots, sentence_{ji} \rbrace$ and the word set belonging to the sentence $s_{ij}$ is further decomposed into $\lbrace word_{1ji}, word_{2ji}, word_{3ji}, \dots ,word_{kji} \rbrace$. Our segmentd corpus provides an explicit multi-level representation of textual data tailored for downstream numerical processing in the next step.

\begin{figure}[htbp]
		\centering
		\includegraphics[width=0.96\textwidth, keepaspectratio]{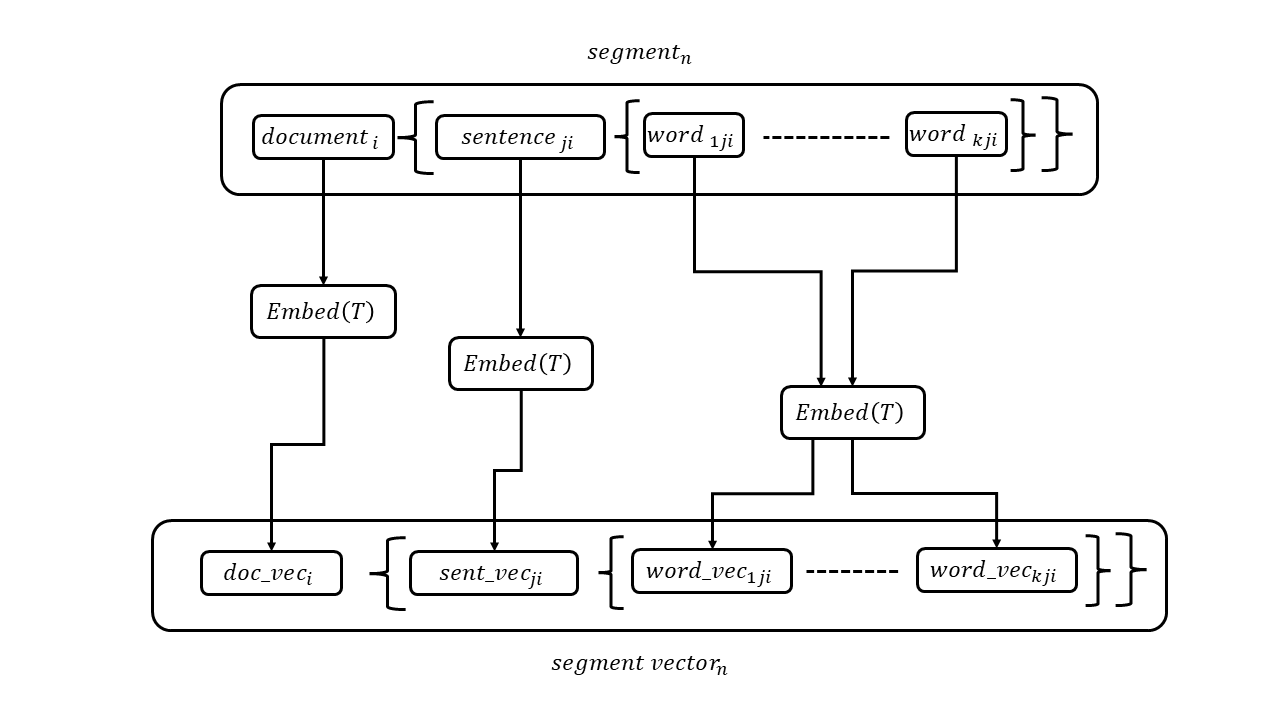}
		\caption{Multi-level vectorization of a segment}
		\label{fig:multilevelvectorization}
	\end{figure}

	\paragraph{Step 2: Multi-level Vectorization}
	\label{paragraph:multi-level vectorization}
	To encode textual data, we designed a multi-level vectorization strategy to embed the nested components we made in~\ref{paragraph:multi-level segmentation}. Specifically, we applied a \emph{layer averaging} strategy to the BERT pretrained model~\citep{bert}, enabling us to extract contextual and positional representations of our data from the model. In transformer-based language models, each input token is represented by a hidden state vector at every layer of the model. These hidden states capture different levels of linguistic and semantic information as the input passes through the layers. The upper layers of these embeddings produced more context-specific representations than the lower layers. Less than 5\% of the variance on average, in contextualized representations, might be defined by static embedding~\citep{bertlayers}. Eq.~\ref{eq:layeraveraging}, formulates how we average the hidden states of several upper transformer layers of BERT. Formally, for an input textual unit $T$, we define the embedding function as:
	
	\begin{equation}
		Embed(T) = \frac{1}{L} \sum_{l=l_1}^{l_2} \mathbf{h}^{(l)}   (T),
		\label{eq:layeraveraging}
	\end{equation}
	
	where $\mathbf{h}^{(l)} (T)$ is the hidden state from layer $(l)$ for the input $T$, and $ l_1, l_2 $ are the lowest and highest target layers of the model, which makes $L$ the number of layers aggregated respectively. As demonstraited in Eq.~\ref{eq:segcorp}, we segmented each sample of our corpus into a multi-level segment, and now we feed units of the segments to the $Embed(T)$ function. Figure~\ref{fig:multilevelvectorization} illustraits how a segment is vectorized in a multi-level manner.
	
	Aggregating hidden states across multiple transformer layers is a technique to enhance the quality of contextual embeddings; this approach allows the model to fuse information from different depths, where lower layers typically capture syntactic and lexical features, and higher layers encode more abstract and semantic patterns; by averaging the hidden states over a range of layers, the resulting representation benefits from richer linguistic features and reduced sensitivity to noise from any single layer; aggregating the top layers of transformer models often enhances performance on downstream tasks, including text classification, semantic similarity, and representation learning~\citep{ma2019, eindor2020}. This layer aggregation strategy produces a comprehensive and semantically meaningful embedding for the input text, which is the core of our multi-level vectorization technique.
	
\begin{figure}[htbp]
		\centering
		\includegraphics[width=\textwidth]{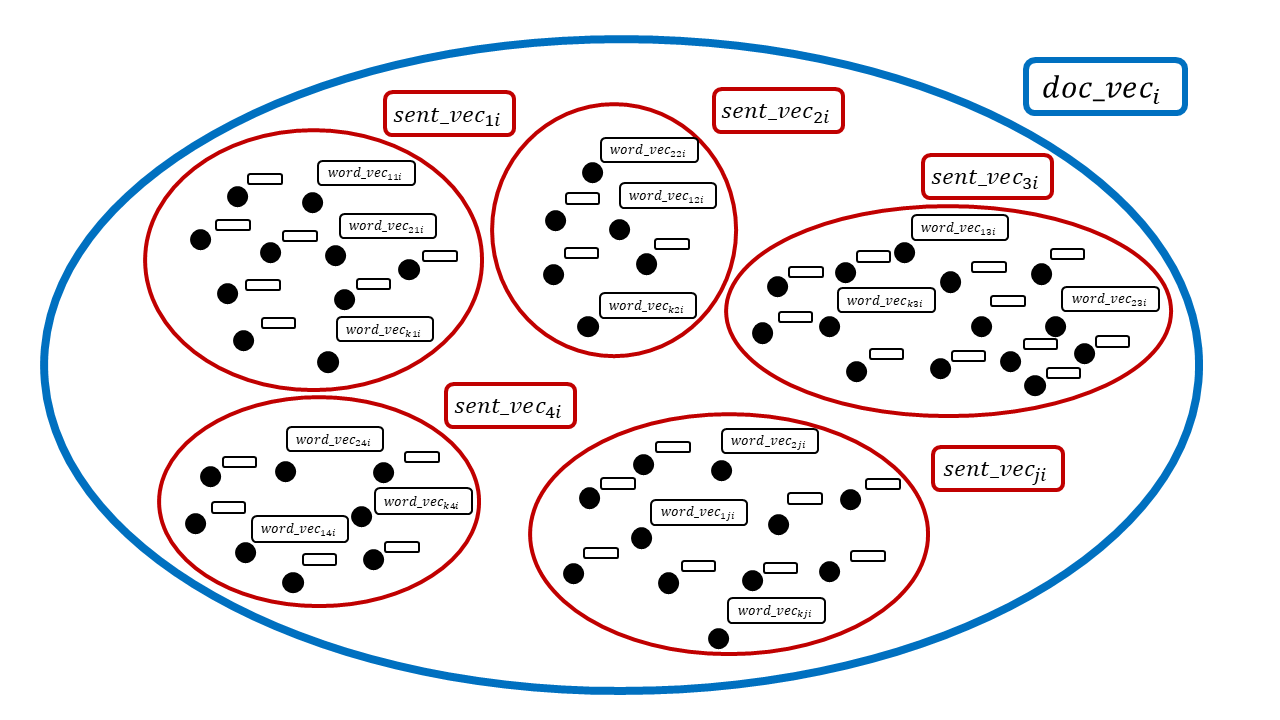}
		\caption{Sample illustration of a hypergraph constructed from a document}
		\label{fig:hyperg}
	\end{figure}

	\paragraph{Step 3: Hypergraph Construction}
	\label{paragraph:hypergraph-construction}
	To capture the hierarchical structure of text, we construct a hypergraph $H = (V, E)$ for each document, to explicitly model the hierarchical structure of text in a multi-level manner. Each the document ($D$) and its sentences ($E = E^{(s)} \cup D$) form hyperedges, while words ($V = \lbrace w_j \rbrace$) serve as the fundamental nodes. Specifically, each sentence hyperedge encompasses its constituent word nodes ($S_j \subseteq V$) , and the document hyperedge embraces all its sentences ($D = \bigcup_{j} S_{j}$). Each node (word) and each hyperedge (sentences and document) is associated with an embedding vector derived from BERT and stored in nested structures shown in Figure~\ref{fig:hyperg}, capturing both intra-level (sentence to words) and inter-level (document to sentences) hierarchies, enabling the expressive modeling of semantic relationships across different textual levels.

\begin{figure}[htbp]
		\centering
		\includegraphics[width=\textwidth]{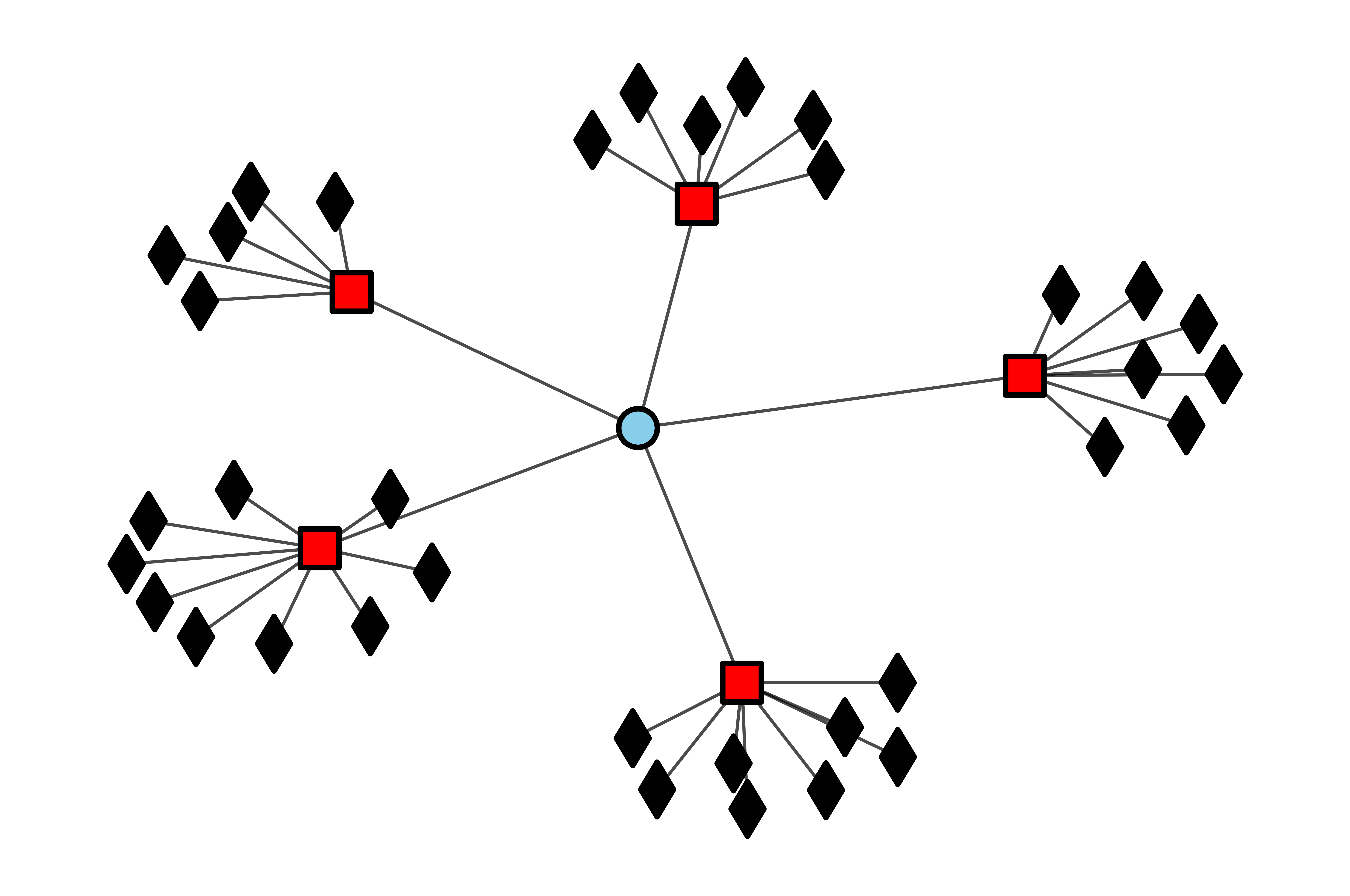}
		\caption{Sample illustration of a hierarchical graph contructed from a hypergraph}
		\label{fig:hirarg}
	\end{figure}

	\paragraph{Step 4: Hypergraph to Hierarchical Graph Transformation}
	\label{paragraph:hypergraph to hierarchical graph}
	To overcome the constraint of pairwise relations, which is the default assumption of graph neural networks, while preserving the hypergraph's semantic hierarchy, we transform each hypergraph to a hierarchical graph structure (Figure~\ref{fig:hirarg}). Firstly, the document node (circle node in Figure~\ref{fig:hirarg}) is placed in the core, representing the document hyperedge, and is equipped with its corresponding embedding vector as a node feature. Then the sentence nodes (square nodes in Figure~\ref{fig:hirarg}) are attached to the document node with their corresponding embedding vectors as node features. The word nodes (rhombus nodes in Figure~\ref{fig:hirarg}) are placed and attached to their parent sentences afterwards, and the corresponding embedding vectors are added as their node feature. Eq.~\ref{eq:cosine}, show the how the edge weight $e_{n_i, n_j}$ between nodes $n_i$ and $n_j$ is set to cosine similarity additional non-negativity constraint: 
	\begin{equation}
		e_{n_i, n_j} = \max(0, \frac{{v_{n_i}}^\intercal v_{n_j}}{\Vert v_{n_i} \Vert\, \Vert v_{n_j} \Vert}),
		\label{eq:cosine}
	\end{equation}
	where each node is associated with a feature vector $v_{n_i}$. The resulting weight vetor is bouned in $\left[ 0, 1 \right]$ ensuring that all edges represent non-negative, similarity-based affinities, guiding the transformer-based graph encoder introduce in the next phase to focus on linguistically relevant interactions. 
	
	\subsubsection{Phase 3: Representation Learning}
	\label{paragraph:fatureselection}
\begin{figure}[htbp]
		\centering
		\includegraphics[width=0.96\textwidth, keepaspectratio]{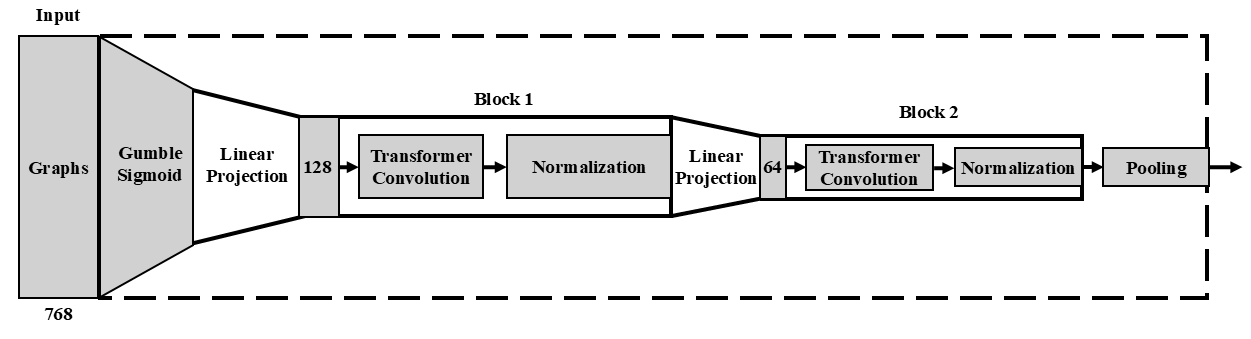}
		\caption{Architecture of the Transformer-Based Graph Encoder (representation learning-phase 3)}
		\label{fig:transformer-encoder}
	\end{figure}

	In this section, we introduce a learnable feature selection mechanism at the input stage of this phase to identify the most informative node attributes for downstream personality prediction. The architecture of tihs phase is depicted in Figure \ref{fig:transformer-encoder}. A learnable weight vevtor $n \in \mathbb{R}^{d}$, where $d$ is the dimention of node features of node $n$, generates a differentiable hard feature mask using the gumbel-sigmoid strategy. Eq.~\ref{eq:feature-mask} demonstrates resulting binary mask $\mathcal{M}$, which stochastically selects a subset of features, effectively suppressing irrelevant dimensions while remaining fully differentiable:
	
	\label{subsubsec:rep-learning-dim-reduction}
	
	\begin{equation}
		\mathcal{M} = \text{GumbleSigmoid}(\mathbf{n} / \tau),
		\label{eq:feature-mask}
	\end{equation}
	where $\tau$ is the temperature parameter controlling the \textbf{sharpness} of the selection. The masked input is obtained in Eq.~\ref{eq:elementwisemultip}:
	\begin{equation}
		\tilde{\mathbf{x}}_i = \mathbf{x}_i \odot \mathbf{m},
		\label{eq:elementwisemultip}
	\end{equation}
	where $\odot$ denotes element-wise multiplication. This adaptive feature selection reduces noise and promotes sparsity in the learned representations. The gumbel-sigmoid is a continuous relaxation of discrete Bernoulli sampling, enabling stochastic binary decisions to remain differentiable and thus trainable with backpropagation. It leverages the gumbel-max technique, which samples from a categorical distribution by perturbing logits with Gumbel noise and selecting the maximum. For the binary case, this mechanism is adapted into a sigmoid form. Concretely, for each feature weight $ s_j $ in the learnable vector $\mathbf{s}$, according to Eq.~\ref{eq:gumbel-sigmoid}, the gumbel-sigmoid generates a sample:
	
	\begin{equation}
		m_j = \sigma\left(\frac{s_j + g_j}{\tau}\right), \quad g_j \sim \text{Gumbel}(0,1),
		\label{eq:gumbel-sigmoid}
	\end{equation}
	where $ g_j $ is a random variable sampled from the standard Gumbel distribution, $ \tau $ is the temperature parameter, and $ \sigma(\cdot) $ denotes the logistic sigmoid function. The temperature $ \tau $ controls the \textbf{smoothness }of the approximation: as $ \tau \to 0 $, the distribution becomes nearly discrete, producing binary-like selections that approximate hard masking, whereas higher $ \tau $ values yield softer, probabilistic masks. This property allows the model to transition from exploration (soft, noisy selections at higher $ \tau $) to exploitation (sharp, deterministic selections at lower $ \tau $) during training. As shown in, the Gumbel-Sigmoid trick preserves differentiability despite producing near-binary masks. Gradients flow through the continuous relaxation, allowing the feature selection mask $\mathcal{M}$ to be optimized jointly with the graph encoder via standard backpropagation, enabling end-to-end learning of which input features are most relevant while suppressing noisy or redundant dimensions without requiring explicit non-differentiable thresholding.
	
	The backbone of the encoder is a transformer-based graph operator that performs attention-driven message passing over graph nodes, as shown in Eq.~\ref{eq:updated-layer} the $TransformerConv$ operator leverages self-attention to adaptively weight neighbor contributions based on their contextual relevance. This design enables the encoder to integrate both local structural dependencies and semantic correlations among connected nodes. At layer $l$, the hidden representation of node $i$ is updated as:
	
	\begin{equation}
		\mathbf{h}_i^{(l)} = \textrm{Sigmoid}\Bigg(\textrm{LayerNorm}\Big( \mathbf{W}_{\text{res}}^{(l)} \mathbf{h}_i^{(l-1)} 
		+ \textrm{TransformerConv}^{(l)}(\mathbf{h}_i^{(l-1)}) \Big)\Bigg),
		\label{eq:updated-layer}
	\end{equation}
	where $\mathbf{h}_i^{(0)} = \tilde{\mathbf{x}}_i$ corresponds to the masked input after feature selection, and $\mathbf{W}_{\text{res}}^{(l)}$ is a residual projection matrix that aligns dimensions between successive layers, ensuring stable gradient flow. The use of residual connections mitigates vanishing gradients, allowing the model to preserve essential node-level knowledge across deeper layers. The LayerNorm term standardizes feature statistics within each node embedding, thereby reducing covariate shift and accelerating training convergence. Finally, a sigmoid nonlinearity is applied to bound the latent representations into a normalized range, which helps stabilize subsequent pooling and classification. The core operator, $\textrm{TransformerConv}^{(l)}$, extends the message-passing paradigm with a scaled dot-product attention mechanism. Eq.~\ref{eq:transformerconv-update} shows the update rule is for node $i$ as:
	
	\begin{equation}
		\mathbf{x}'_i = \mathbf{W}_1 \mathbf{x}_i + \sum_{j \in \mathcal{N}(i)} \alpha_{i,j} \mathbf{W}_2 \mathbf{x}_j,
		\label{eq:transformerconv-update}
	\end{equation}
	where $\mathcal{N}(i)$ denotes the neighborhood of node $i$. The first term $\mathbf{W}_1 \mathbf{x}_i$ preserves self-information through a learnable transformation, while the second term aggregates neighbor information, modulated by the attention coefficient $\alpha_{i,j}$. This coefficient reflects the importance of neighbor $j$ in updating node $i$, and is computed via multi-head scaled dot-product attention in Eq.~\ref{eq:attention-coeff}:
	
	\begin{equation}
		\alpha_{i,j} = \text{softmax}\left(\frac{(\mathbf{W}_3 \mathbf{x}_i)^\top (\mathbf{W}_4 \mathbf{x}_j)}{\sqrt{d}}\right),
		\label{eq:attention-coeff}
	\end{equation}
	here, $\mathbf{W}_3$ and $\mathbf{W}_4$ project node features into query and key spaces, respectively, while the denominator $\sqrt{d}$ normalizes the inner product to avoid sharp gradients in high-dimensional settings. The softmax function ensures that the attention coefficients form a probability distribution over the neighbors of $i$, thereby enforcing relative weighting of structural dependencies. Multi-head attention can be incorporated by repeating this computation across several independent subspaces and either concatenating or averaging the results. However, in our implementation, we use a single head to maintain computational efficiency and interpretability. Intuitively, this mechanism allows each node to selectively focus on the most semantically aligned neighbors, rather than treating all neighbors as equally informative. Unlike traditional GCN-style averaging, the TransformerConv operator dynamically adapts its receptive field based on both node content and graph structure, which are particularly important in text-derived graphs, where syntactic neighbors may carry varying degrees of importance depending on context. As a result, the encoder can integrate hierarchical linguistic cues while simultaneously suppressing noisy or weakly correlated signals.
	
	After two transformer blocks, the node embeddings capture both local structural dependencies and attention-weighted contextual features. To obtain a holistic representation of the entire graph, we employ a global additive pooling operation shown in Eq.~\ref{eq:graph-embedding}, aggregating the final node embeddings into a fixed-dimensional vector:
	
	\begin{equation}
		\mathbf{z}_G = \sum_{i \in G} \mathbf{h}_i^{(2)},
		\label{eq:graph-embedding}
	\end{equation}
	where $\mathbf{h}_i^{(2)}$ denotes the final embedding of node $i$ after the second encoder block. This summation is permutation-invariant, meaning that the representation remains unaffected by the ordering of nodes in the input graph; a property essential for processing arbitrary text-derived graphs, where node order has no intrinsic meaning.  
	
	The choice of additive pooling is motivated by its ability to preserve cumulative contributions from all nodes, thereby capturing the overall distribution of linguistic and structural cues present in the graph and allowing highly informative nodes to exert proportionally, being particularly beneficial in the context of personality prediction, where specific nodes (words or sentences with strong psychological markers) should contribute more prominently to the final representation. Formally, additive pooling can be interpreted as learning a bag-of-embeddings representation of the graph, where the magnitude of $\mathbf{z}_G$ reflects both the strength and frequency of node-level knowledge. This aggregated vector thus encodes both feature information (via learned node embeddings) and structural information (via attention-driven message passing). The resulting graph-level embedding $\mathbf{z}_G$ serves as the input to the personality classification head, where it is mapped to predictions of the Big Five trait scores.

	\subsubsection{Phase 4: Personality Classification}
	\label{subsubsec:personality-classification}
	\begin{figure}[b]
		\centering
		\includegraphics[width=\textwidth, keepaspectratio]{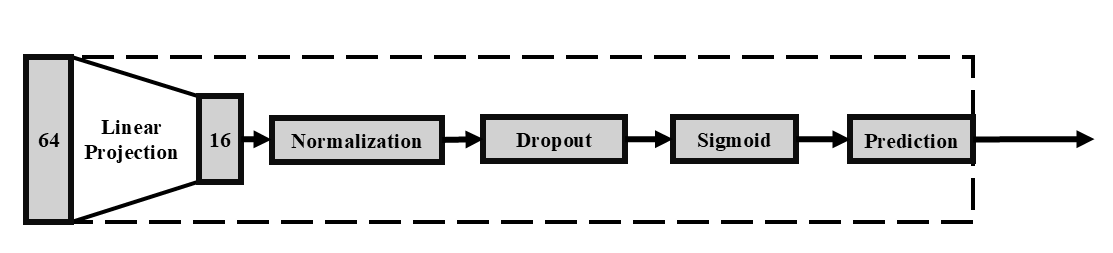}
		\caption{Architecture of the personality classification with sigmoid-gated latent representation}
		\label{fig:personality-classification}
	\end{figure}
	
	As illustrated in Figure~\ref{fig:personality-classification}, the personality classification maps the pooled graph embedding $\mathbf{z}_G \in \mathbb{R}^{ld}$, where $ld$ is a compact latent space before producing label prediction. The classification head is designed to be lightweight yet expressive, ensuring that the personality prediction is driven primarily by the structural and semantic representations learned in the graph encoder while still benefiting from regularization and non-linear transformations. Eq.~\ref{eq:classification}, demonstraits the forward pass through the classification proceeds as follows:
	
	\begin{equation}
		\begin{aligned}
			\mathbf{z}_1 &= \mathbf{W}_1 \mathbf{z}_G + \mathbf{b}_1, \quad \mathbf{z}_1 \in \mathbb{R}^{ld}, \\
			\mathbf{z}_2 &= \textrm{LayerNorm}(\mathbf{z}_1), \\
			\mathbf{z}_3 &= \textrm{Dropout}(\mathbf{z}_2), \\
			\mathbf{z}_4 &= \sigma(\mathbf{z}_3), \\
			\hat{\mathbf{y}} &= \mathbf{W}_2 \mathbf{z}_4 + \mathbf{b}_2, \quad \hat{\mathbf{y}} \in \mathbb{R}^{1},
		\end{aligned}
		\label{eq:classification}
	\end{equation}
	where $\mathbf{W}_1, \mathbf{W}_2$ and $\mathbf{b}_1, \mathbf{b}_2$ are learnable projection parameters. The intermediate latent activation $\mathbf{z}_4$ acts as a sigmoid-gated bottleneck representation, while $\hat{\mathbf{y}}$ corresponds to the raw logits for the five personality traits.
	
	Layer normalization stabilizes hidden activations and prevents internal covariate shift, which is crucial in small latent spaces where fluctuations in scale can disproportionately affect prediction quality. The subsequent dropout layer provides regularization by randomly setting elements of the normalized vector to zero, thereby reducing overfitting and encouraging the model to rely on distributed representations rather than memorizing false correlations. The inclusion of a sigmoid nonlinearity ensures that the latent features are bounded between $[0,1]$, which can be interpreted as gating the latent representation before projection to the trait space. This design introduces an additional layer of nonlinearity, enhancing the classifier's expressive power without incurring significant computational overhead. The final linear projection maps the gated latent representation into the output space of the label. Since personality prediction is inherently a binary classification problem, Eq.~\ref{eq:bceloss} shows how the network is optimized using the binary cross-entropy (BCE) loss:
	
	\begin{equation}
		\mathcal{L}_{\text{BCE}} = - \frac{1}{N} \sum_{k=1}^{N} \Big( y_k \log \sigma(\hat{y}_k) + (1-y_k) \log(1-\sigma(\hat{y}_k)) \Big),
		\label{eq:bceloss}
	\end{equation}
	where $ y_k \in \{0,1\} $ denotes the ground-truth binary label for the $k^{\text{th}}$ sample, and $ \hat{y}_k \in \mathbb{R} $ represents the corresponding raw model output (logit). The logistic sigmoid function is defined in Eq.~\ref{eq:logisticsigmoid}:
	
	\begin{equation}
		\sigma(\hat{y}_k) = \frac{1}{1 + e^{-\hat{y}_k}},
		\label{eq:logisticsigmoid}
	\end{equation}
	mapping logits to the continuous interval $(0,1)$, allowing it to be interpreted as the probability of the positive class. The binary cross-entropy (BCE) objective measures the dissimilarity between the predicted probability $ \sigma(\hat{y}_k) $ and the true label $ y_k $, penalizing predictions that deviate from the target. The first term, $ y_k \log \sigma(\hat{y}_k) $, drives the model to assign high probabilities to positive instances ($ y_k = 1 $), while the second term, $ (1 - y_k) \log(1 - \sigma(\hat{y}_k)) $, discourages assigning high probabilities to negative instances ($ y_k = 0 $). The summation averages these contributions over all $ N $ samples in the batch, providing a stable estimate of the expected prediction error. The sigmoid operation is typically applied internally during loss computation, combining the sigmoid and logarithmic operations into a single, numerically stable step that prevents underflow or overflow. Consequently, the network outputs raw logits $ \hat{y}_k $ and is optimized end-to-end to jointly predict binary labels.
	
	By combining compact linear projections, normalization, dropout regularization, and a sigmoid-gated latent representation, the classifier effectively bridges the high-dimensional graph embeddings with the interpretable predictions. This modular design ensures that the graph encoder handles the bulk of representational learning, while the classifier enforces robustness and facilitates efficient binary label mapping.
	
	\subsection*{Code availability}
	\label{subsec:codeavailability}
	
	The source code used to implement the proposed HyperPersona framework and reproduce the experiments is publicly available at \url{https://github.com/SinaHeydari76/HyperPersona}.
	
	\subsection*{Declaration of Generative AI and AI-Assisted Technologies}
	\label{subsec:aiassistdeclaration}
	
	During the preparation of this manuscript, generative AI tools were used only to a limited extent for language editing, including grammar and spelling correction, phrasing, and improving the clarity and readability of the text. No AI tools were used to develop the scientific methodology, conduct experiments, analyze or generate data, interpret results, or formulate the conclusions. All scientific content was developed, critically reviewed, and verified by the authors, who take full responsibility for the accuracy and integrity of the final manuscript.
	
	\section{Results \& Discussion}
	\label{sec:results}
	
	\subsection{Evaluation Metrics}
	\label{subsec:evaluationmetrics}
	
	To comprehensively assess the performance of the binary classification model, we use four standard and widely adopted metrics for evaluating APP performance including: \emph{accuracy}, \emph{precision}, \emph{recall}, and \emph{f1-score} \citep{ramezani2022, kazameini2020, ramezani2022oe, wang2020, el2022}. These measures collectively capture the global correctness of predictions, providing a robust characterization of model behavior under potential class imbalance.
	All metrics are computed based on the confusion matrix components (true positives (TP), false positives (FP), false negatives (FN), and true negatives (TN)). Accuracy (Eq.~\ref{eq:accuracy}) quantifies the overall proportion of correctly classified instances, regardless of class. 
	
	\begin{equation}
		Accuracy = \frac{TP + TN}{TP + FP + TN + FN}.
		\label{eq:accuracy}
	\end{equation}
	
	Accuracy provides an intuitive summary of performance by indicating the fraction of correct predictions across all samples.
	However, in imbalanced datasets where one class is significantly more frequent than the other, accuracy alone can be misleading, as a model may achieve high accuracy simply by favoring the majority class.
	Consequently, accuracy should be interpreted alongside class-sensitive metrics such as f1-score. The f1-score (Eq.~\ref{eq:f1score}) provides a harmonic mean between precision and recall, serving as a balanced indicator of performance when both metrics are important.  
	
	\begin{equation}
		F1-score = 2 \times \frac{Precision \times Recall}{Precision + Recall}.
		\label{eq:f1score}
	\end{equation} 
	
	F1-score, is an informative measure that summarizes the trade-off between false positives (FP) and false negatives (FN). In practice, the f1-score is particularly useful for evaluating models under class imbalance, where accuracy may fail to reflect true discriminative performance. Unlike the arithmetic mean, the harmonic mean penalizes large disparities between precision and recall. The precision (Eq.~\ref{eq:precision}) evaluates the reliability of the model’s positive predictions. 
	It measures the proportion of correctly identified positive instances among all samples predicted as positive. High precision indicates that when the model predicts the positive class, it is usually correct despite producing few false positives.  
	This property makes precision especially valuable in contexts where false alarms carry a high cost. Nevertheless, precision alone does not account for missed positive instances, which are instead captured by recall. The recall (Eq.~\ref{eq:recall}), also known as sensitivity or the true positive rate, measures the model’s ability to correctly identify all positive instances. A high recall value indicates that the model successfully detects most of the true positives, with few false negatives. Recall is particularly critical in applications where missing a positive instance is more costly than incorrectly predicting one.
	
	\begin{equation}
		Precision = \frac{TP}{TP + FP},
		\label{eq:precision}
	\end{equation}
	\begin{equation}
		Recall = \frac{TP}{TP + FN}.
		\label{eq:recall}
	\end{equation}
	
	\subsection{Parameter Settings}
	\label{subsec:parametersetting}
	\begin{table}[b]
		\centering
		\caption{Parameter settings of the Transformer-Based Graph Encoder on the Essays dataset.}
		\label{tab:parameter}
		\begin{tabular}{ll}
			\toprule
			\textbf{Parameter} & \textbf{Value} \\
			\midrule
			Train-test split ratio (\%) & 80 - 20 \\
			Training epochs & 30 \\
			Batch size & 8 \\
			Optimizer & AdamW \\
			Learning rate & $3 \times 10^{-4}$ \\
			Weight decay & $3 \times 10^{-4}$ \\
			Dropout rate & 0.2 \\
			Loss function & BCEWithLogitsLoss \\
			\bottomrule
		\end{tabular}
	\end{table}
	This section details the hyperparameters and implementation settings used in the experiments. As presented in Table~\ref{tab:parameter}, the preprocessing and multi-level segmentation were performed using spaCy~\citep{spacy}. Graph and hypergraph related operations were conducted with NetworkX~\citep{networkx} and HyperNetX~\citep{hypernetx}, respectively. Primary features were represented using pre-computed embeddings extracted from BERT~\citep{bert}. As illustrated in Figure~\ref{fig:transformer-encoder}, the proposed transformer-based graph encoder was implemented using PyTorch~\citep{pytorch} and PyTorch Geometric~\citep{pgeometric}. We used two TransformerConv layers to project node features into a 128 and 64-dimensional latent space, followed by layer normalization and residual linear projection. Ultimately, we performed global additive pooling to derive a graph-level embedding in a 64-dimensional feature space. As demonstrated in Figure~\ref{fig:personality-classification}, the~\nameref{subsubsec:personality-classification} is a classification head consisting of a linear layer (64→16), layer normalization, sigmoid activation, dropout, and a final linear layer to produce the output logits.
	
	\subsection{Baseline Methods}
	\label{subsec:baselinemethods}
	
	To assess the effectiveness of \emph{HyperPersona}, it is crucial to benchmark its performance against well-established baseline methods in text-based personality prediction. These baselines represent a spectrum of methodological approaches, ranging from traditional machine learning and ontological reasoning to modern deep learning architectures leveraging contextual embeddings. Each method provides unique insights into how textual features can be harnessed for personality inference, and therefore serves as a meaningful point of comparison.
	
	\begin{itemize}
		\item Tighe et al.~\cite{tighe2016} employed LIWC features to conduct APP on the Essays dataset using several classifiers, including SVM, Sequential Minimal Optimization, and Linear Logistic Regression. They focused on enhancing classification performance by eliminating insignificant LIWC features through Information Gain and Principal Component Analysis (PCA).
		\item Majumder et al.~\cite{majumder2017} proposed a convolutional neural network architecture to infer an author’s Big Five personality traits from text. They built one binary classifier per trait, all sharing a common architecture, in which sentence-level convolutional layers encode features that are then aggregated to document-level representations.
		\item Ramezani et al.~\cite{ramezani2022oe} integrated lexical, semantic, and sequential levels. Their ensemble combined term-frequency vectorization, ontology-based and enriched ontology models, latent semantic analysis, and BiLSTM-based deep learning. These base models were stacked into a meta-model using a Hierarchical Attention Network, which performed reasoning across levels.
		\item Wang et al.~\cite{wang2020} proposed a graph convolutional neural network that represents users and their textual information within a graph structure, with the resulting embeddings classified via a softmax layer.
		\item Kazameini et al.~\cite{kazameini2020} combined contextual BERT embeddings with an ensemble of bagged SVMs. Their approach preserved token-level representations and aggregated them through bootstrap sampling, effectively bridging contextual embeddings with robust ensemble learning.
		\item El-Demerdash et al.~\cite{el2022} proposed the use of Universal Language Model Fine-Tuning (ULM-FiT) for author personality prediction (APP), an effective transfer learning approach applicable across various natural language processing tasks.
		\item Jiang et al.~\cite{jiang2020} leveraged pre-trained contextual embeddings, specifically BERT and RoBERTa, to enhance the accuracy of predictions in author personality prediction (APP).
	\end{itemize}
	
	\subsection{Evaluation Results}
	\label{subsec:experimentalresultsanddiscussions}
	
\begin{figure}[htbp]
		\centering
		\includegraphics[width=\textwidth, keepaspectratio]{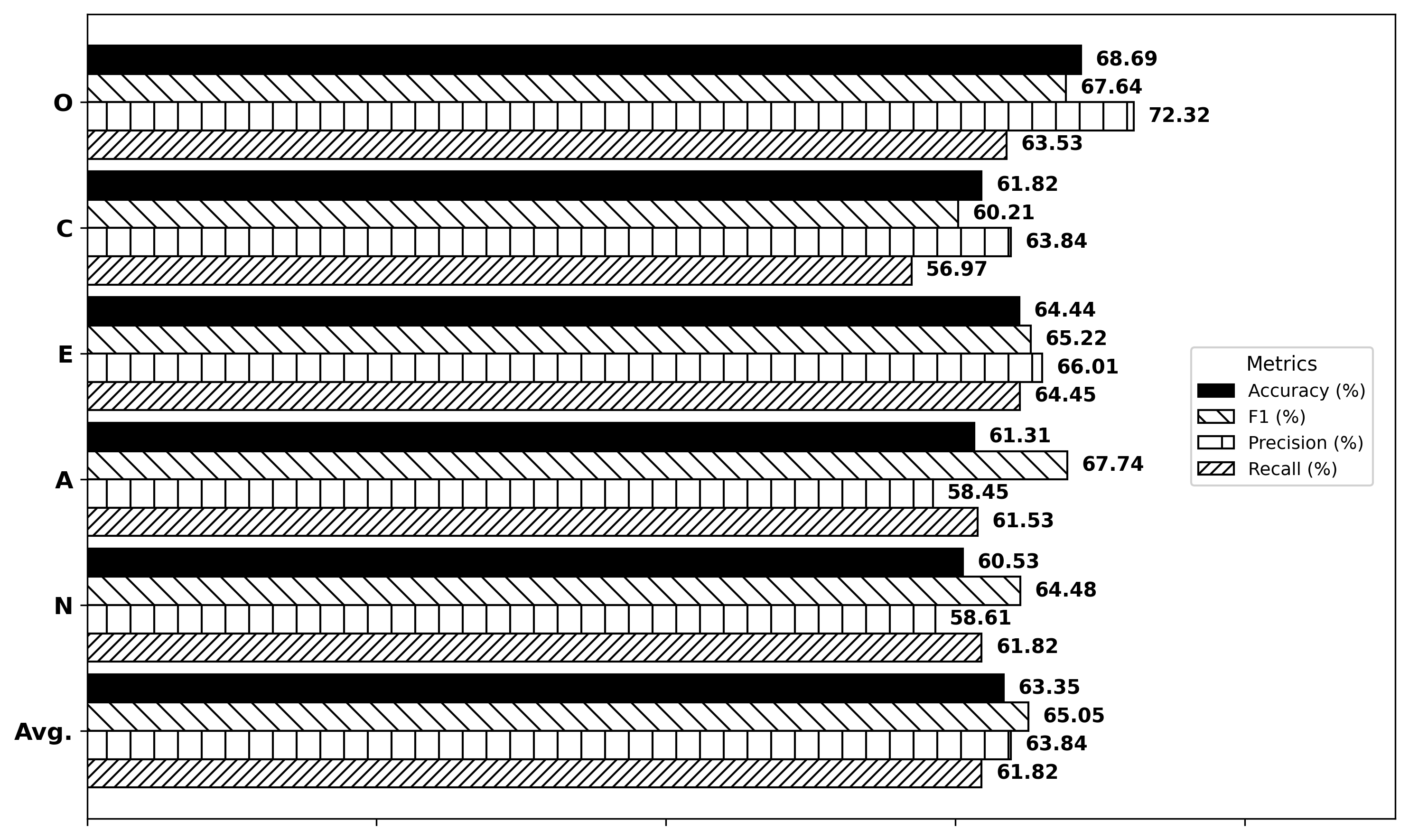}
		\caption{Performance of the HyperPersona across the Big Five personality traits, evaluated using accuracy, f1-score, precision, and recall metrics.}
		\label{fig:modelperformance}
	\end{figure}

	Table~\ref{tab:results} provides a detailed per-trait evaluation of our model across standard performance metrics. Openness exhibits the highest overall performance, indicating that the model is most effective in predicting this trait. Extraversion also yields strong and balanced results across all metrics, suggesting consistent performance. Conscientiousness, Agreeableness, and Neuroticism show slightly lower performance, reflecting the greater challenge of inferring these traits from text. The average accuracy across all traits is 63.15\%, with an f1-score of 65.05\%. Additionally, the model achieves 63.82\% precision and 61.82\% recall, demonstrating its ability to make reliable and consistent predictions. These results demonstrate that HyperPersona delivers stable and competitive performance across all traits, with particularly strong predictive capabilities. Figure~\ref{fig:modelperformance} presents a comparative visualization of the model’s performance across the five personality traits, underscoring the model’s effectiveness in multi-trait prediction, with powerful generalization for cognitively and socially expressive traits.
	
	\begin{table}[htbp]
		\centering
		\caption{HyperPersona (proposed method) evaluation results on Essays Dataset per Big Five traits} 
		\label{tab:results}
		\begin{tabular}{lllll}
			\toprule 
			\textbf{Personality Trait} & \textbf{Accuracy(\%)} & \textbf{F1(\%)} & \textbf{Precision(\%)} & \textbf{Recall(\%)} \\
			\midrule
			\textbf{O} & 68.69 & 67.64 & 72.32 & 63.53 \\
			\textbf{C} & 61.82 & 60.21 & 63.84 & 56.97 \\
			\textbf{E} & 64.44 & 65.22 & 66.01 & 64.45 \\
			\textbf{A} & 61.31 & 67.74 & 58.45 & 61.53 \\
			\textbf{N} & 60.53 & 64.48 & 58.61 & 61.82 \\
			\midrule
			\textbf{Average} & \textbf{63.35} & \textbf{65.05} & \textbf{63.84} & \textbf{61.82}\\
			\bottomrule
	\end{tabular}\end{table}
	
	\begin{figure}[htbp]
		\centering
		\begin{subfigure}[b]{0.45\textwidth}
			\includegraphics[width=\textwidth]{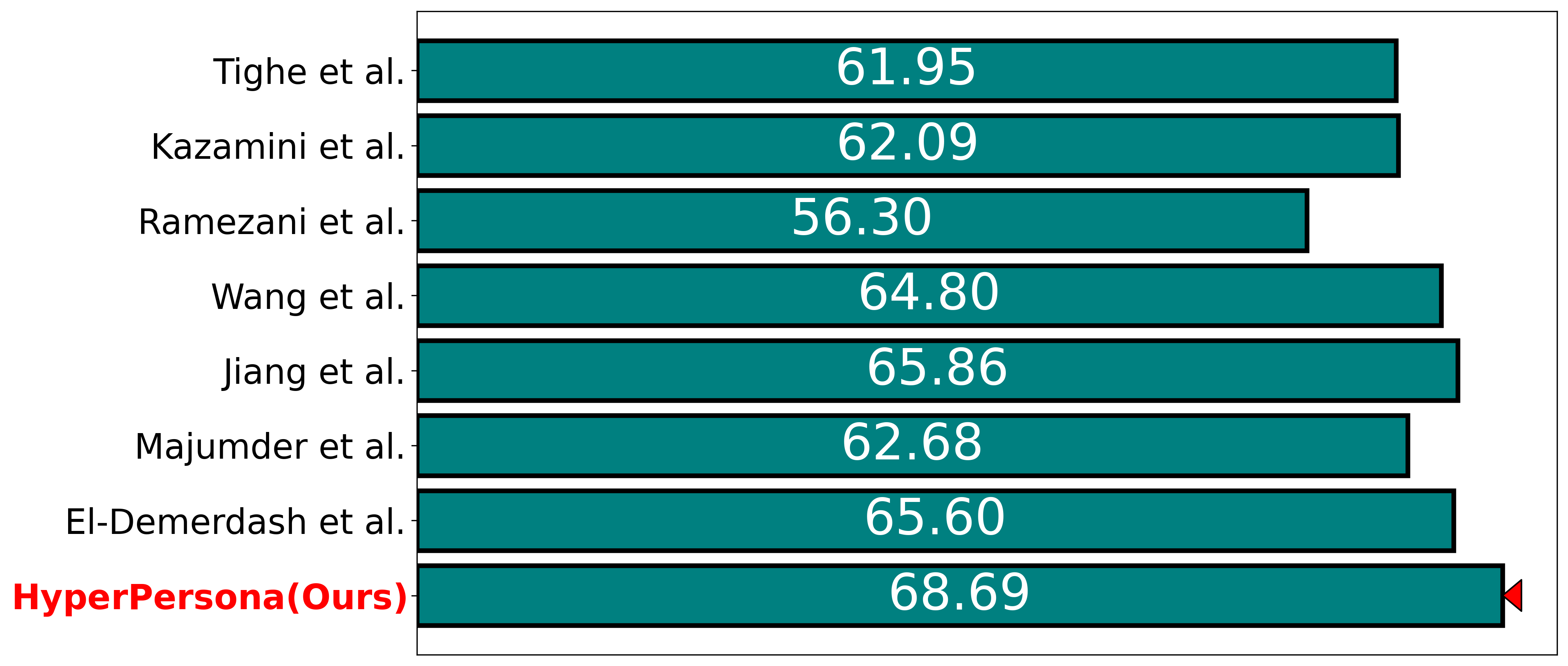}
			\caption{O}
			\label{subfig:acc-openness}
		\end{subfigure}
		\hfill
		\begin{subfigure}[b]{0.45\textwidth}
			\includegraphics[width=\textwidth]{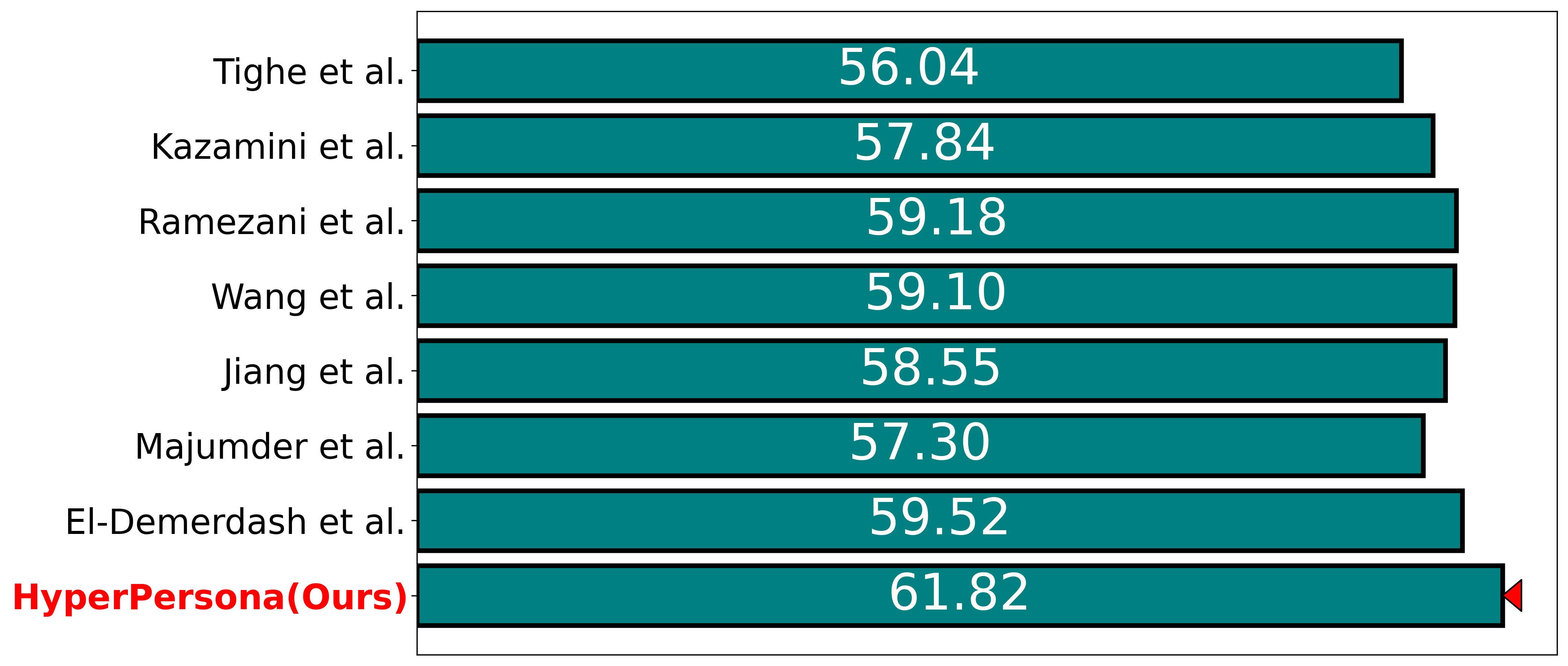}
			\caption{C}
			\label{subfig:acc-conscientiousness}
		\end{subfigure}
		\hfill
		\begin{subfigure}[b]{0.45\textwidth}
			\includegraphics[width=\textwidth]{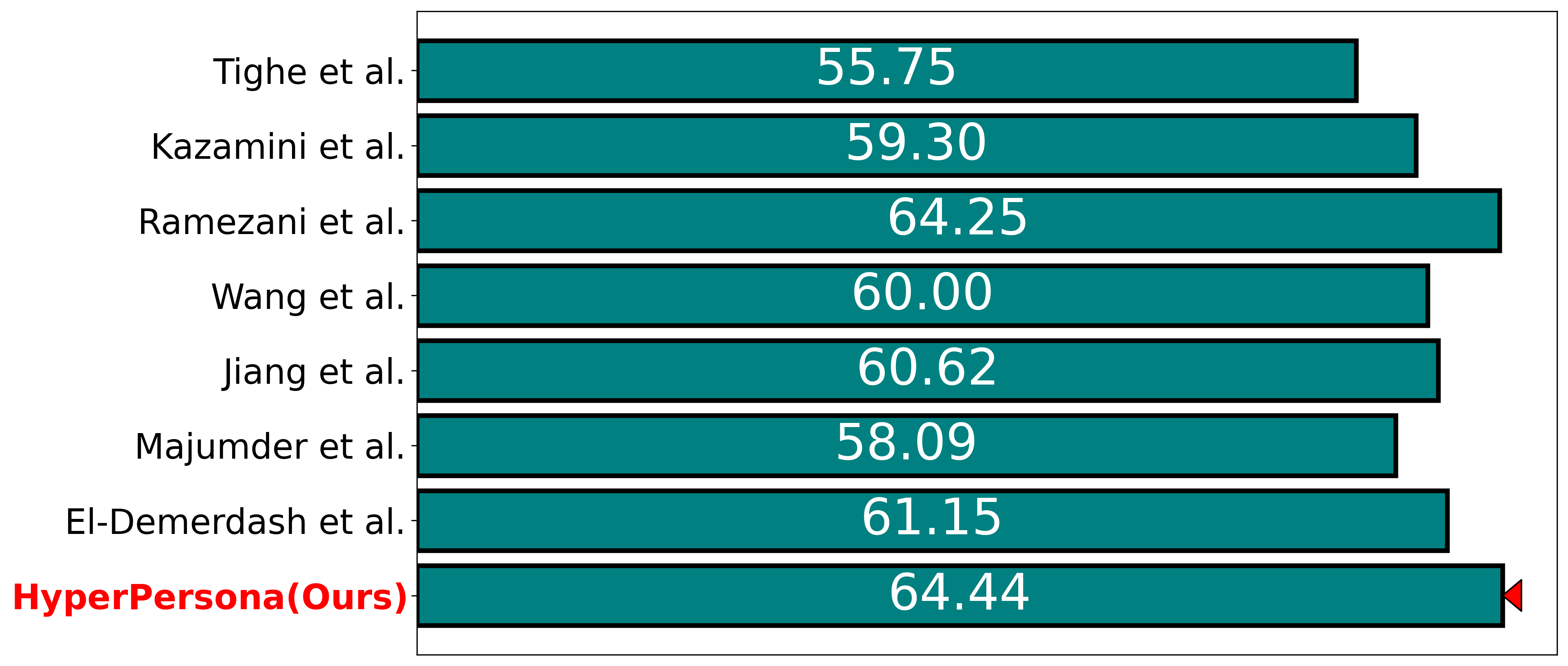}
			\caption{E}
			\label{subfig:acc-extraversion}
		\end{subfigure}
		\hfill
		\begin{subfigure}[b]{0.45\textwidth}
			\includegraphics[width=\textwidth]{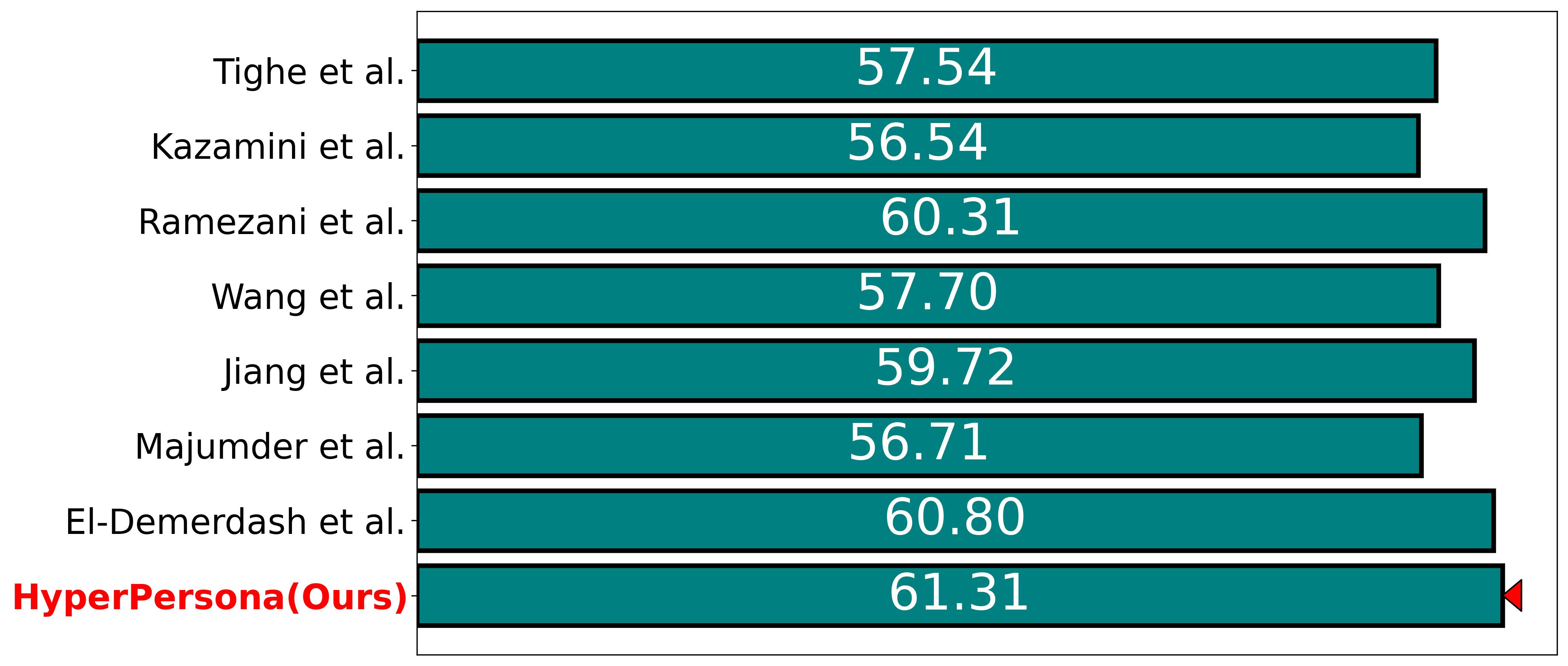}
			\caption{A}
			\label{subfig:acc-agreeableness}
		\end{subfigure}
		\hfill
		\begin{subfigure}[b]{0.45\textwidth}
			\includegraphics[width=\textwidth]{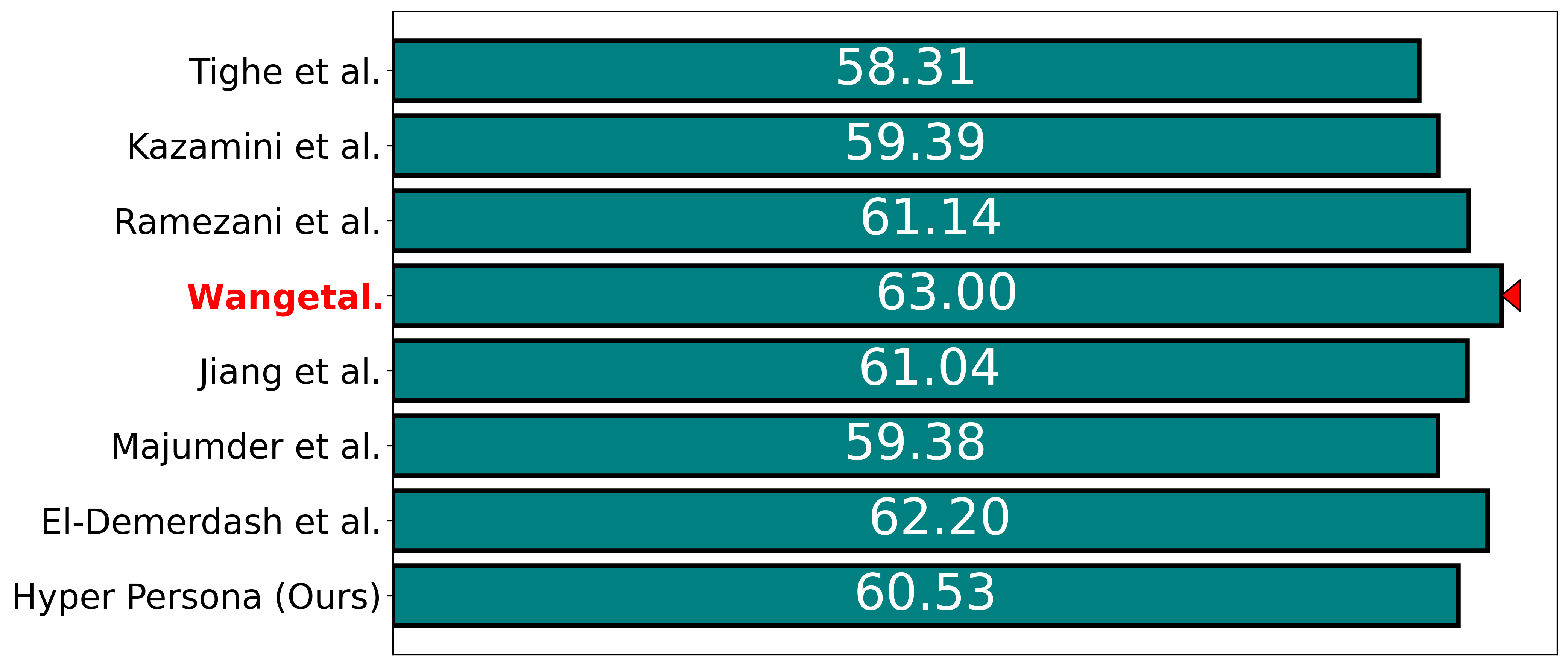}
			\caption{N}
			\label{subfig:acc-neuroticism}
		\end{subfigure}
		\hfill
		\begin{subfigure}[b]{0.45\textwidth}
			\includegraphics[width=\textwidth]{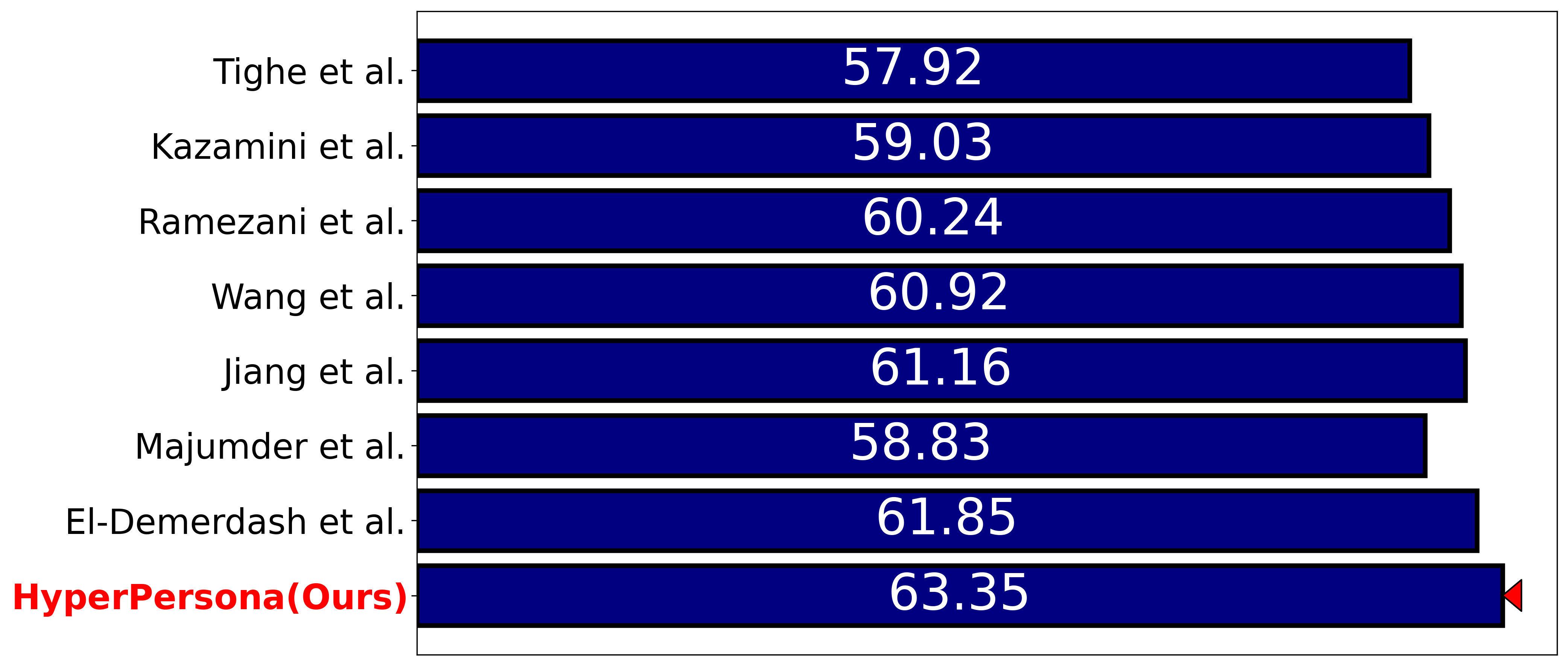}
			\caption{Avg.}
			\label{subfig:acc-average}
		\end{subfigure}
		\caption{Comparative prediction accuracy results per personality trait and overall average of HyperPersona against baseline methods.}
		\label{fig:accuracy}
	\end{figure}
	
	\begin{table}[htbp]
		\centering
		\caption{Comparing the prediction accuracy of HyperPersona against baseline methods.}
		\label{tab:compare}
		\begin{tabular}{lllllll}
			\toprule
			\textbf{Baselines}&\multicolumn{6}{c}{\textbf{Accuracy (\%)}}\\
			\cmidrule{2-7}
			&\textbf{O} & \textbf{C} & \textbf{E} & \textbf{A} & \textbf{N} & \textbf{Avg.}\\
			\midrule
			Tighe et al.~\cite{tighe2016} & 61.95 & 56.04 & 55.75 & 57.54 & 58.31 & 57.92 \\
			Kazameini et al.~\cite{kazameini2020} & 62.09 & 57.84 & 59.30 & 56.54 & 59.39 & 59.03 \\
			Ramezani et al.~\cite{ramezani2022oe} & 56.30 & 59.18 & 64.25 &  60.31 & 61.14 & 60.24 \\
			Wang et al.~\cite{wang2020} & 64.80 & 59.10 & 60.00 & 57.70 & \textbf{63.00} & 60.92 \\ 
			Jiang et al.~\cite{jiang2020} &  65.86 & 58.55 & 60.62 & 59.72 & 61.04 & 61.16 \\
			Majumder et al.~\cite{majumder2017} & 62.68 & 57.30 & 58.09 & 56.71 & 59.38 & 58.83 \\
			El-Demerdash et al.~\cite{el2022} & 65.60 & 59.52 & 61.15 & 60.80 & 62.20 & 61.85 \\
			\midrule
			\textbf{HyperPersona (Ours)} & \textbf{68.69} & \textbf{61.82} & \textbf{64.44} & \textbf{61.31} & 60.53 & \textbf{63.35} \\
			\bottomrule
		\end{tabular}
	\end{table}

	\begin{figure}[htbp]
		\centering
		\includegraphics[width=\textwidth, keepaspectratio]{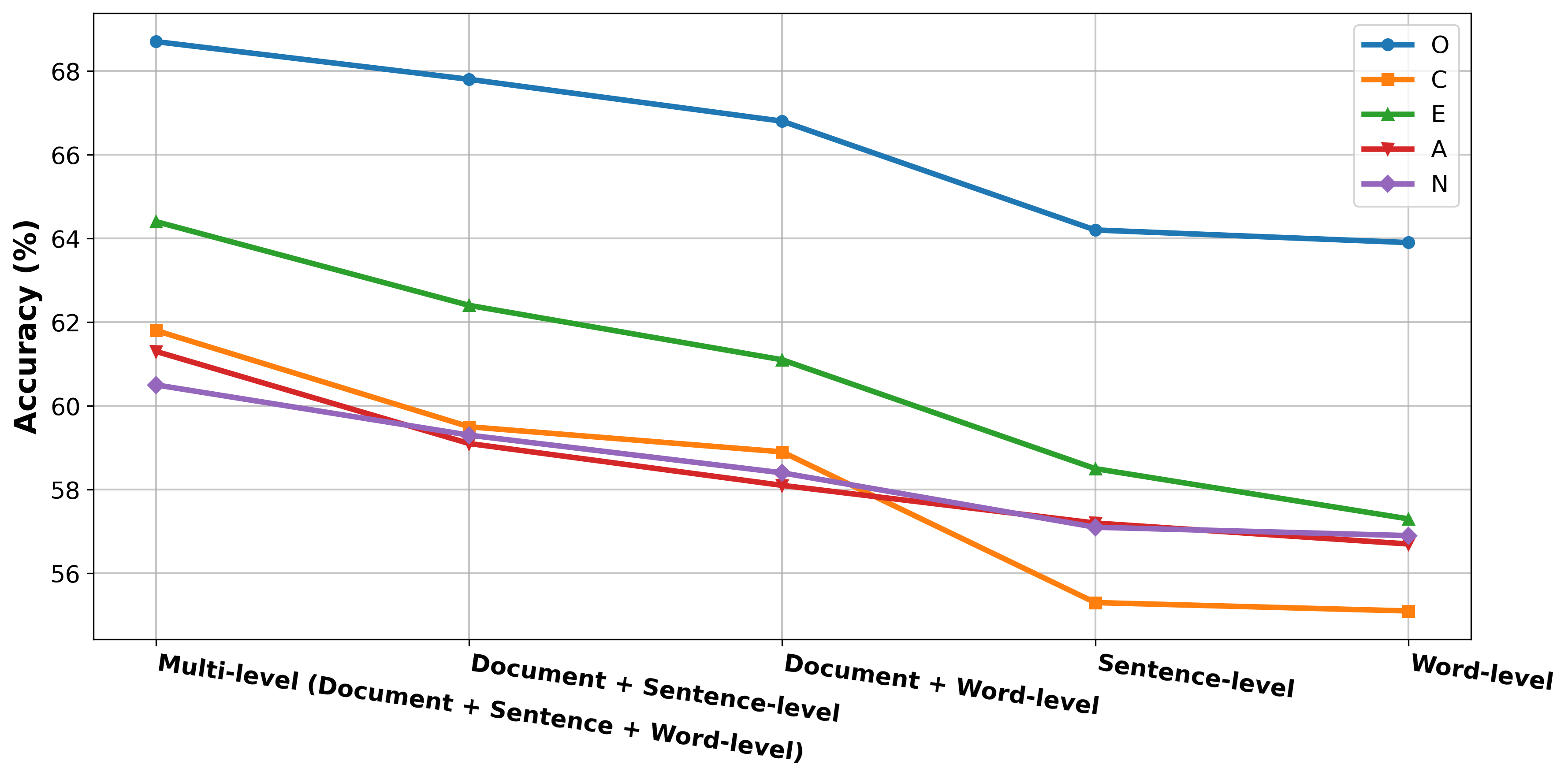}
		\caption{Visulization of ablation study across different representation level combiantion for each Big Five personality trait.}
		\label{fig:replevelsperformance}
	\end{figure}
	
	\begin{sidewaystable}[htbp]
		\centering
		\caption{Ablation study across different representation level combiantion for each Big Five personality trait.}
		\label{tab:replevelsperformance}
		\begin{tabular}{p{7.5cm} llllll}
			\toprule
			\textbf{Text Level(s)}&\multicolumn{6}{c}{\textbf{Accuracy (\%)}}\\
			\cmidrule{2-7}
			& \textbf{O} & \textbf{C} & \textbf{E} & \textbf{A} & \textbf{N} & \textbf{Avg.} \\
			\midrule
			\textbf{Multi-level (Document + Sentence + Word-level)} & \textbf{68.69} & \textbf{61.82} & \textbf{64.44} & \textbf{61.31} & \textbf{60.53} & \textbf{63.35} \\
			Document + Sentence-level & 67.81 & 59.51 & 62.35 & 59.11 & 59.31 & 61.61 \\
			Document + Word-level & 66.80 & 58.88 & 61.15 & 58.10 & 58.97 & 60.66 \\
			Sentence-level & 64.19 & 55.31 & 58.55 & 57.16 & 57.12 & 58.46 \\
			Word-level & 63.82 & 55.08 & 57.22 & 56.73 & 56.94 & 57.95 \\
			\bottomrule
		\end{tabular}
	\end{sidewaystable}
	
	Table~\ref{tab:compare} presents a comparative analysis against baseline methods. Baseline methods, achieved moderate performance levels, reflecting the inherent difficulty of modeling personality traits solely from textual data, particularly within the Essays dataset. Ontology-based and semantically enriched architectures, such as those presented by Ramezani et al.~\cite{ramezani2022oe}, have achieved modest improvements by leveraging structured lexical and semantic information. In contrast, HyperPersona surpasses all baseline methods across three personality traits and in overall performance, highlighting its ability to extract richer contextual cues from textual input. Although the model yields slightly lower results in agreeableness and neuroticism, this can be attributed to the subtler and less distinctive linguistic markers associated with these traits within the dataset. The overall results demonstrate that HyperPersona offers a substantial and stable advancement over existing methods, effectively capturing both expressive and cognitive language patterns.
	
	Figure~\ref{fig:accuracy} illustrates the comparative classification accuracy across the Big Five personality dimensions, along with the overall average. The results confirm that the proposed \textit{HyperPersona} model consistently outperforms prior approaches, achieving the highest average accuracy. This consistent improvement highlights the model’s capability to effectively capture intricate linguistic and contextual cues relevant to personality expression. The results establish that HyperPersona consistently exceeds existing methods in predictive accuracy. However, beyond numerical gains, it is crucial to understand how the model’s structural design contributes to these improvements. The following discussion interprets these findings in light of the research questions and the theoretical motivation underlying multi-level text modeling.
	
	\subsection{Discussion}
	\label{sec:discussion}
	
	This study demonstrates that representing text as interconnected hypergraphs significantly enhances the ability to infer personality traits from written language alone. By integrating document, sentence, and word-level representations through a transformer-based graph encoder, HyperPersona captures hierarchical dependencies that conventional models flatten. These findings suggest that personality cues are distributed across linguistic layers and that hypergraph modeling offers an effective computational means to unify them.
	
	\begin{itemize}
		\item	Addressing \textbf{RQ1}, the proposed multi-level hypergraph structure proves crucial for enriching the semantic and syntactic representation of text. By jointly modeling word, sentence, and document-level information and their interactions, the framework captures both local lexical associations and global contextual dependencies. Unlike traditional flat or sequential representations, it encodes relationships among linguistic units at multiple granularities, enabling the model to preserve discourse coherence while retaining fine-grained syntactic and semantic nuances. This integration yields a unified representation that bridges micro-level lexical patterns and macro-level thematic organization, providing a richer foundation for personality prediction.
		\item	In relation to \textbf{RQ2}, the comparative results illustrated in Figure~\ref{fig:accuracy} and Table~\ref{tab:compare} reveal that the proposed HyperPersona framework consistently surpasses established baseline methods across most of the Big Five traits. HyperPersona’s capacity to jointly encode multi-level structural dependencies enables it to extract richer personality cues from text than single-level or sequential approaches. Its superiority is particularly evident for openness, conscientiousness, extraversion, and agreeableness, where nuanced linguistic variety and social tone are more effectively captured through hierarchical modeling. Even for traits such as Neuroticism, where personality expressions are linguistically subtle, the model maintains stable and competitive performance, highlighting its robustness in handling diverse textual styles and personality manifestations.
		\item	Moving to \textbf{RQ3}, Figure~\ref{fig:replevelsperformance} and Table~\ref{tab:replevelsperformance} reveal an evident performance decline as the representation shifts from the complete multi-level structure to abblated-level variants. As shown in Figure~\ref{fig:replevelsperformance} the trend quantifies the contribution of each abstraction layer: document-level features provide global thematic context, sentence-level connections encode syntactic and discourse structure, and word-level cues capture lexical and affective knowledge. The synergy among these layers yields the most accurate representations of personality. In contrast, when individual levels are isolated (such as in sentence or word-only configurations), the model loses access to important contextual dependencies, leading to notable performance degradation. These results affirm the necessity of multi-granular fusion for robust and comprehensive personality inference.
		\item	Finally, addressing \textbf{RQ4}, the results presented in Figure~\ref{fig:modelperformance} and Table~\ref{tab:results} demonstrate that the transformer operating over the hypergraph effectively aggregates personality-relevant linguistic knowledge across multiple levels of abstraction. Through the attention mechanism, the model selectively emphasizes high-impact nodes, such as semantically salient words or sentences that convey personality cues (for instance, emotional intensity for Neuroticism or social expressiveness for Extraversion). This hierarchical propagation enhances both the interpretability and the sensitivity of the learned embeddings, demonstrating that the architecture generalizes personality indicators across different linguistic layers while maintaining consistent performance across the Big Five personality traits.
	\end{itemize}
	\section{Conclusion}
	\label{sec:conclusion}
	
	This study was designed to examine the impact of multi-level text representations using hypergraphs and hierarchical graph structures, and transformer-based graph encoders on text-based Automatic Personality Prediction (APP). To this end, we proposed \emph{HyperPersona}, a framework that leverages hypergraph and hierarchical graph structures to capture the multi-level nature of written text (document, sentence and word) and employs a transformer-based graph encoder for feature learning. \emph{HyperPersona} operates through four main phases. First, a preprocessing phase prepares the input text for subsequent processing. Second, a text representation strategy transforms each sample to a digestable form for machine. Third, representation learning and dimensionality reduction are performed using a transformer-based graph encoder. Finally, learned representations are classified using a multi layer perceptron.To the best of our knowledge, this is the first work to employ hypergraph and hierarchical graph structures for text representation to perform text-based APP. The findings from this study offer several noteworthy contributions to the existing literature. First, the proposed multi-level approach significantly improves the accuracy of text-based APP systems that rely solely on written text. Second, embedding text using pre-trained transformers provides contextual awareness and eliminates the need for traditional preprocessing techniques. Third, this study advances our understanding of how hypergraphs and hierarchical graphs can effectively exploit textual hierarchy, offering a more comprehensive and structured representation of written text compared to conventional flat sequences of characters highlighting the importance of the proposed multi-level representation learning in enabling machines to perform more human-like decision-making.

	\bibliographystyle{unsrtnat}
	\bibliography{references} 
	
\end{document}